\documentclass[10pt,conference]{IEEEtran}
\IEEEoverridecommandlockouts
\usepackage{cite}
\usepackage{amsmath,amssymb,amsfonts}
\usepackage{booktabs}
\usepackage{multirow}
\usepackage{ragged2e}
\usepackage{caption}
\usepackage{adjustbox}
\usepackage{tikz}
\usepackage{float}
\usepackage{graphicx}

\usepackage{hyperref}
\usepackage{enumerate}
\usepackage{subfig}

\usepackage{pgfplots}

\usepackage{textcomp}
\usepackage{xcolor}
\usepackage{booktabs}

\usepackage{pgfplots}
\usepackage{tikz}
\usetikzlibrary{spy, backgrounds}
\usepackage{color, colortbl}

\newcommand{\fref}[1]{Fig.~\ref{#1}}

\usepackage[misc]{ifsym}

\usepackage{textcomp}

\usepackage[utf8]{inputenc}
\usepackage{enumitem}

\usepackage{tikz}
\newcommand*\circled[1]{\tikz[baseline=(char.base)]{
            \node[shape=circle,draw,inner sep=0.1pt] (char) {#1};}}
\newcounter{stepnum}
\newcommand{\steplabel}{\bfseries Step \circled{\arabic{stepnum}}\stepcounter{stepnum}}

\usepackage[outline]{contour}
\contourlength{0.2pt}
\contournumber{15}

\newcommand\HUGE{\fontsize{22.7}{25}\selectfont}

\def\BibTeX{{\rm B\kern-.05em{\sc i\kern-.025em b}\kern-.08em
    T\kern-.1667em\lower.7ex\hbox{E}\kern-.125emX}}

\usepackage[linesnumbered,ruled,vlined]{algorithm2e}

\SetKwComment{Comment}{$\triangleright$\ }{}
\SetKwInput{kwInit}{Input}
\SetKwInput{KwResult}{Output}

 \usepackage{fancyhdr}
\fancypagestyle{mystyle}{
    \chead{\large The 2024 IEEE International Conference on Big Data (IEEE BigData 2024)}}
\begin{document}


%



\title{\HUGE {LPLgrad: Optimizing Active Learning Through Gradient Norm Sample Selection and Auxiliary Model Training}}

\author{\IEEEauthorblockN{
Shreen Gul\IEEEauthorrefmark{2}, Mohamed Elmahallawy\IEEEauthorrefmark{3}, Sanjay Madria\IEEEauthorrefmark{2}, Ardhendu Tripathy\IEEEauthorrefmark{2}}  
      \IEEEauthorblockA{%
 \IEEEauthorrefmark{2}Computer Science Department, Missouri University of Science and Technology, Rolla, MO 65401, USA}
       \IEEEauthorblockA{%
 \IEEEauthorrefmark{3} School of Engineering \& Applied Sciences, Washington State University, Richland, WA 99354, USA}
Emails:  sgchr@mst.edu, mohamed.elmahallawy@wsu.edu, madrias@mst.edu, astripathy@mst.edu

\thanks{This work was supported by the National Science Foundation (NSF) under Grant No. 2246187.} 
}

\maketitle
\thispagestyle{mystyle}

\begin{abstract}

Machine learning models are increasingly being utilized across various fields and tasks due to their outstanding performance and strong generalization capabilities. Nonetheless, their success hinges on the availability of large volumes of annotated data, the creation of which is often labor-intensive, time-consuming, and expensive. Many active learning (AL) approaches have been proposed to address these challenges, but they often fail to fully leverage the information from the core phases of AL, such as training on the labeled set and querying new unlabeled samples. To bridge this gap, we propose a novel AL approach, Loss Prediction Loss with Gradient Norm (LPLgrad), designed to quantify model uncertainty effectively and improve the accuracy of image classification tasks. LPLgrad operates in two distinct phases: (i) {\em Training Phase} aims to predict the loss for input features by jointly training a main model and an auxiliary model. Both models are trained on the labeled data to maximize the efficiency of the learning process—an aspect often overlooked in previous AL methods. This dual-model approach enhances the ability to extract complex input features and learn intrinsic patterns from the data effectively; (ii) {\em Querying Phase} that quantifies the uncertainty of the main model to guide sample selection. This is achieved by calculating the gradient norm of the entropy values for samples in the unlabeled dataset. Samples with the highest gradient norms are prioritized for labeling and subsequently added to the labeled set, improving the model's performance with minimal labeling effort.  Extensive evaluations on real-world datasets demonstrate that the LPLgrad approach outperforms state-of-the-art methods by order of magnitude in terms of accuracy on a small number of labeled images, yet achieving comparable training and querying times in multiple image classification tasks.  Our code is available at \href{https://github.com/sgchr273/LPLgrad}{Github}.

\end{abstract}


\section{Introduction}\label{sec:intro}

Machine learning models are being adopted rapidly across various fields due to their exceptional performance and generalization capabilities. These models rely on both data and ground-truth labels to excel in their tasks. However, obtaining ground-truth labels is often challenging. For instance, in medical imaging, domain experts must be compensated to annotate data points, and in speech recognition, labeling audio data at the word level can take significantly more time than the actual speech duration\cite{zhan2022comparative}. The manual annotation process is both time-consuming and labor-intensive\cite{li2024survey,gul2024fishermask}.

\begin{figure}[!t]
    \centering
    \includegraphics[width=0.5\textwidth]{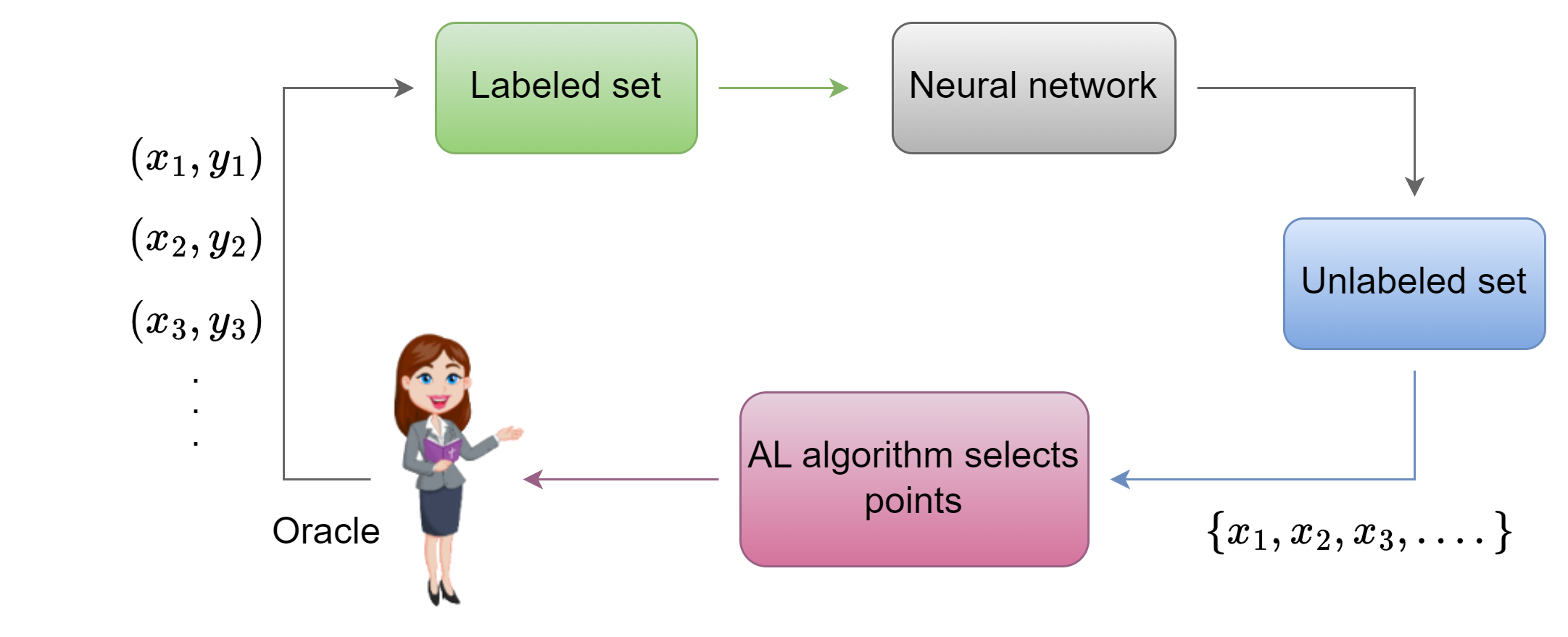}
    \caption{Typical active learning pipeline.\vspace{-0.5cm}} 
    \label{fig:AL}
\end{figure}

Active learning (AL) offers a solution to these challenges by intelligently selecting the most informative data points for labeling, thereby reducing the overall annotation effort \cite{gilhuber2022verips}. In each active learning round, a set of new unlabeled points is selected for annotation and added to the labeled set, and then the target model will be trained on this updated labeled set (\fref{fig:AL} presents an overview of a typical AL process). Various AL methodologies have been proposed in recent years, generally categorized into {\em uncertainty sampling} and {\em diversity sampling}. Uncertainty sampling targets data points where the model is most uncertain about their categories, with methods such as entropy sampling \cite{wang2014new}, margin sampling \cite{netzer2011reading}, and least-confidence sampling \cite{wang2014new} being popular examples. Diversity sampling, on the other hand, aims to select the most diverse samples that represent the entire dataset, with recent approaches including Coreset \cite{sener2017active}, variational adversarial AL (VAAL) \cite{sinha2019variational}, and Wasserstein
adversarial AL (WAAL) \cite{shui2020deep}. 

\begin{figure*}
    \centering
    \includegraphics[width=1\textwidth]{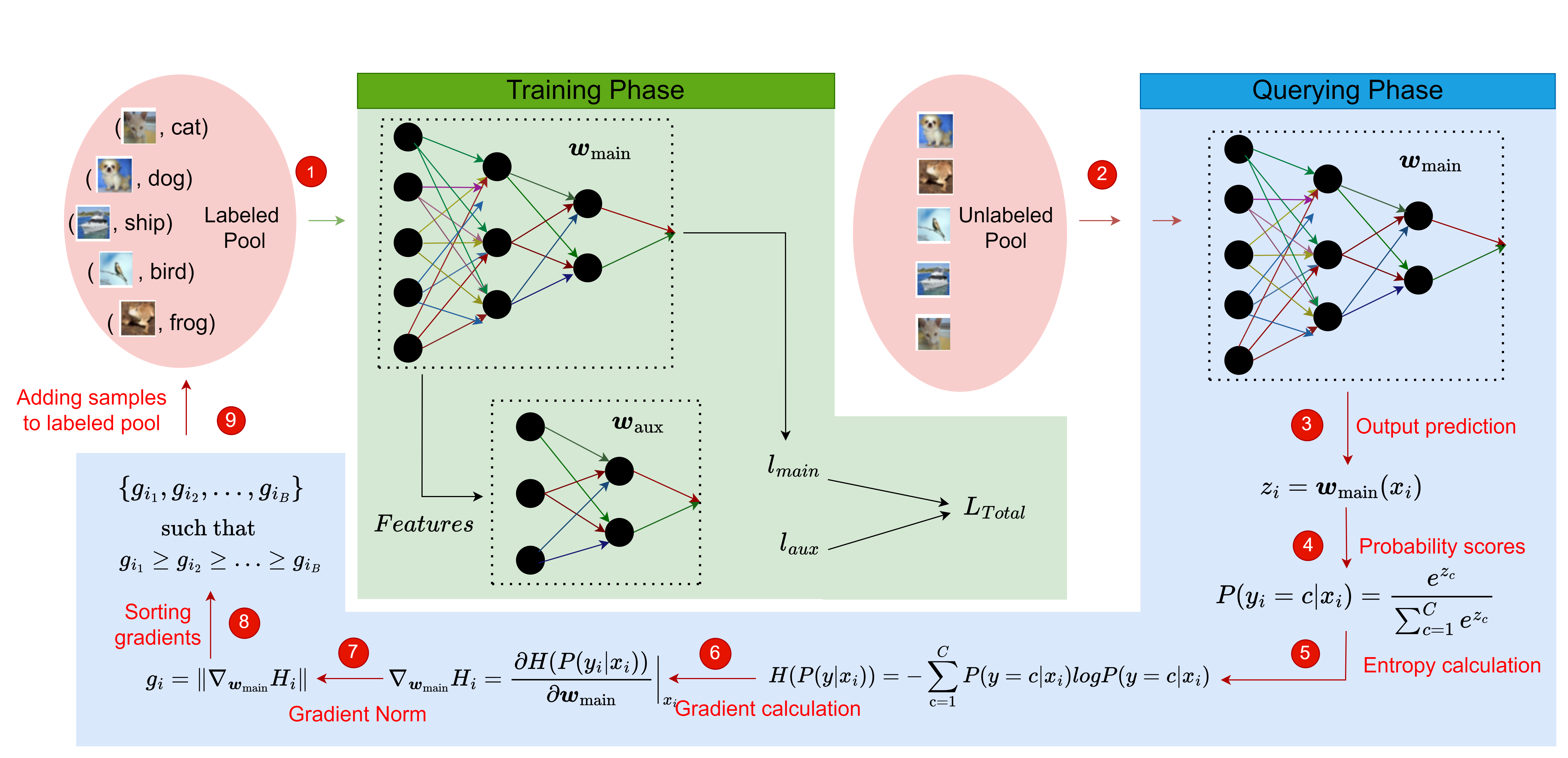}
  \caption{A visualization of our LPLgrad approach. It highlights the two key components of LPLgrad: training and querying phases. The process involves 9 steps (represented in red circles): In steps 1 and 2, labeled images \(\mathcal{L}\) are processed through the main model \(\boldsymbol{w}_{\text{main}}\), where feature maps are extracted and subsequently fed into the auxiliary model \(\boldsymbol{w}_{\text{aux}}\). This combination of models yields two distinct losses, \(l_{\text{main}}\) and \(l_{\text{aux}}\), which are then aggregated to compute the total loss \(L_{\text{total}}\). From steps 3 to 9, samples from the unlabeled set \(\mathcal{U}\) are processed, producing scores that are passed through a softmax classifier to obtain entropy values for each sample. These entropy values are back-propagated through \(\boldsymbol{w}_{\text{main}}\) to compute the gradient. Then, LPLgrad calculates the gradient norm and selects the samples with the highest gradient norms for annotating and adding to \(\mathcal{L}\).}

    \label{fig:framework}

\end{figure*}

{\bf Challenges.}  Although AL methods reduce traditional manual annotation efforts, they often fail to fully leverage the information from the core phases of AL  such as training on labeled set and querying new unlabeled samples. For example, loss prediction loss (LPL), an uncertainty sampling method, integrates a loss prediction module with the target model, yielding two loss values—target loss and prediction loss—through joint learning. The unlabeled points with the highest prediction loss values are then chosen for labeling\cite{yoo2019learning}. However, the hyperparameters in the loss prediction module can be highly sensitive in large-scale datasets like Tiny ImageNet \cite{le2015tiny} and EMNIST \cite{cohen2017emnist}, leading to performance degradation \cite{zhan2022comparative}. Furthermore, we empirically observe that selecting points based on prediction loss values is {\em less effective} compared to selecting points based on their entropy values (see section~\ref{sec:Exp}).\\
On the other hand, methods that utilize output entropy to calculate gradient norm, as proposed by Wang et al. \cite{wang2022boosting} emphasize the connection between selected samples and model performance on test data to guide sample selection.

{\bf Contributions.} In this paper, we propose a novel AL approach named Loss Prediction Loss with Gradient Norm (LPLgrad). LPLgrad incorporates an entropy-based method to quantify model uncertainty, enhancing the accuracy of image classification. Specifically, our proposed AL approach is inspired by \cite{yoo2019learning} but differs in three key ways: (i) We do not use the loss prediction module during the querying phase as done by \cite{yoo2019learning}; instead, we integrate the loss prediction module as an auxiliary model with the main model only {\em during training phase}  (see the green region in \fref{fig:framework}); (ii) {\em During the querying phase} we utilize network probability scores instead of input features, as they directly indicate the model's confidence in its predictions (see the blue region in \fref{fig:framework}); (iii) We conduct enhanced training by learning the main model and auxiliary model together, and calculate the output entropy and subsequently the gradient norm of the unlabeled instances as measures of uncertainty, instead of predicted loss values as in \cite{yoo2019learning}, to capture the best aspects of both the {\em training and querying phases.}


To sum up, our contributions can be summarized as follows:

 
\begin{itemize}
    \item We propose a novel AL algorithm called LPLgrad that leverages a loss prediction module to learn input data features and effectively quantify the network's uncertainty on unlabeled data based on their gradient norms. To the best of our knowledge, this approach is the first of its kind that addresses a common gap in the literature by utilizing information from {\em both the training and sampling phases} of AL, enabling more informed and deliberate decisions about sample selection.

    \item We integrate a main model throughout the training process with the auxiliary model that predicts loss to simultaneously learn the {\em parametric auxiliary model and main model}. This allows us to effectively extract input features and complex patterns within the data. Subsequently, the model utilizes the gradient norm of unlabeled samples as a metric of uncertainty, aiding in informed decision-making during sample querying.

    \item Extensive evaluation with different annotation budgets on the visual task such as image classification demonstrates the superior performance of the proposed method against the state-of-the-art AL approaches.
\end{itemize}

\section{Related Work}\label{sec:related}
 
In this section, we will review recent works in AL, which can be categorized into three main approaches: uncertainty sampling, diversity sampling and hybrid sampling.

\subsection{Uncertainty based Methods}
This category of AL methods evaluates the {\em informativeness} of unlabeled data points by assessing the uncertainty of the target deep network regarding these points. They prioritize selecting those unlabeled points fro annotation and adding to the labeled set where the model exhibits significant uncertainty. In this context, Wang et al. \cite{wang2014new} introduce a metric for data selection based on model uncertainty, known as entropy sampling. This metric is one of the most widely used for uncertainty quantification and data selection. Some of the recent works that propose active learning approaches include \cite{safaei2024entropic, li2024deep, an2024active}. For instance, the authors of \cite{safaei2024entropic} proposed a technique that incorporates both known and unknown data distributions to measure the model's uncertainty. Another work, \cite{li2024deep} introduced a method that utilizes noise stability in the model's parameters as an uncertainty metric. A recent approach by \cite{an2024active} estimates the model's uncertainty by employing a Gaussian process (GP) model as a surrogate for the baseline neural network learner. Another recent work Verified Pseudo-label Selection for Deep Active Learning (VERIPS) proposed by \cite{gilhuber2022verips} that uses a pseudo-label verification mechanism that consists of a second network only trained on data approved
by the oracle and helps to discard questionable pseudo-labels. 

While the aforementioned methods effectively reduce labeling effort, they share a common drawback: {\em they are susceptible to selecting outliers due to their high uncertainty}. Additionally, focusing predominantly on sampling uncertain points can lead to unreliable model predictions and querying redundant data, ultimately decaying model's performance \cite{li2024survey}.
 
\subsection{Diversity based Methods}

In this category of AL methods, the learner queries examples that are representative of the entire data distribution such as the work in \cite{sener2017active,sinha2019variational}. 
In \cite{sener2017active}, the authors proposed the Coreset approach, which is among the most prominent methods in diversity-based AL. It frames AL as a coreset problem; selecting unlabeled samples based on their geometric properties. Despite its effectiveness, this method is computationally intensive and has delayed sampling times because it requires storing an array of labeled samples for comparison with new samples. The authors of \cite{sinha2019variational} proposed a variational adversarial AL (VAAL)  approach that utilizes a variational autoencoder to learn the distribution of labeled data in latent space, coupled with an adversarial network that discriminates between labeled and unlabeled data. However, VAAL necessitates retraining the VAE multiple times rendering it computationally intensive.


While diversity-based methods effectively capture the underlying data distribution, they may fail to fully leverage the information from unlabeled data necessary for training the task learner. Moreover, these techniques might be insensitive to data points near the decision boundary, even though such points could be crucial for the target model to query \cite{li2024survey}.

In this paper, we propose an uncertainty-based AL algorithm that addresses the limitations above by leveraging both the training and querying phases. Our algorithm effectively learns the features of input data through the joint training of models, thereby extracting inherent patterns in the input points and reducing the likelihood of selecting redundant data. During the selection phase, it chooses unlabeled points based on their {\em gradient norm values}, which provably reduces the test loss.


\section{Proposed Methodology}

In this section, we provide a comprehensive explanation of the components comprising our proposed AL approach, LPLgrad. We start with the problem formulation and an overview of the framework, followed by a detailed description of the two main building blocks of LPLgrad: the Training Phase and the Querying Phase.


\subsection{Problem Formulation and Framework}\label{sec:1}

Given a pool of unlabeled set of data samples denoted as $\mathcal{U} = \{x_i\}_{i=1}^N$ with $N$ is the total number of samples, we aim to solve multi-class classification problem with $C$ categories. To do that, we first construct a labeled set of multi-class $\mathcal{L}_B^t =\{x_i, y_i\}_{i=1}^B$ ($y_i$ represents the label of the data point $x_i$) by randomly selecting $B$ samples from the unlabeled pool $\mathcal{U}^t_{N-B}$. Here, the superscript $t=0,1,\dots$ signifies the current round of AL, which increases by one as the training progresses. We then utilize a set of models $\boldsymbol{w} = \{\boldsymbol{w}_{\text{main}}, \boldsymbol{w}_{\text{aux}}\}$ and train them on $\mathcal{L}^t_B$. The training procedure involves an augmented approach where both the main model $\boldsymbol{w}_{\text{main}}$ and the auxiliary model $\boldsymbol{w}_{\text{aux}}$ are jointly learned.

Once the training on the selected set $\mathcal{L}^t_B$ is completed, we compute the output entropy for all the samples in $\mathcal{U}^t_{N-B}$ (which includes only the remaining unselected samples). These entropy values represent the loss incurred by $\boldsymbol{w}_{\text{main}}$. Subsequently, we update the $\boldsymbol{w}_{\text{main}}$ parameters and store the gradient norm for each sample in the set. These stored values will be then sorted, and a new $B'$ samples with the highest gradient norm will be selected for labeling in the next AL cycle ($t+1$).

In the subsequent cycle, the updated labeled and unlabeled sets are denoted as $\mathcal{L}^{t+1}_{B'}$ and $\mathcal{U}^{t+1}_{N-B'}$, respectively. Then, the main and auxiliary models will be trained on $\mathcal{L}^{t+1}_{B'}$ to update their model weights as ${\boldsymbol{w}_{\text{main}}^{t+1},\boldsymbol{w}_{\text{aux}}^{t+1}}$, respectively. This process of training and querying new samples continues in subsequent AL cycles until a certain accuracy threshold is achieved, a predefined budget of iterations is exhausted, or any other termination criterion.

Below, we provide a detailed description of our proposed LPLgrad framework (see red numbered circles in ~\fref{fig:framework}):

\begin{itemize}[leftmargin=*, label={\steplabel}]

    \item In each AL round $t$, we select a set of labeled images $\mathcal{L}_B^t$ to feed both the main model $\boldsymbol{w}_{\text{main}}$ and the auxiliary model $\boldsymbol{w}_{\text{aux}}$. The main model $\boldsymbol{w}_{\text{main}}$ is designed to extract features from the selected labeled images and then feed these features to the auxiliary model $\boldsymbol{w}_{\text{aux}}$. Subsequently, both models $\boldsymbol{w}_{\text{main}}$ and $\boldsymbol{w}_{\text{aux}}$ produce losses, which are jointly learned to generate the total loss $l_\text{total}=l_\text{main}+l_\text{aux}$.
    
    \item Next, the remaining unlabeled images $\mathcal{U}^t_{N-B}$ will be fed into the trained $\boldsymbol{w}_{\text{main}}$ to calculate the prediction scores.
    
    \item Then, the main model $\boldsymbol{w}_{\text{main}}$ outputs a $1 \times C$ vector of prediction scores, where $C$ is the number of classes in the dataset for our multi-class problem.
    
    \item These output prediction scores are then transformed into a vector of posterior probabilities of the same size using a softmax classifier.

    \item After that, we use these resulting probabilities to compute the entropy (a scalar value) of the model \(\boldsymbol{w}_{\text{main}}\) for each sample \(x_i\) in the unlabeled set \(\mathcal{U}^t_{N-B}\).

    \item  In this step, the entropy value is treated as a loss that the model $\boldsymbol{w}_{\text{main}}$ incurs for that particular sample. Subsequently, the gradient of this loss is calculated with respect to the model's parameters. The size of this gradient corresponds to the number of layers in the model.

    \item Subsequently, the norm of the gradient value computed in the previous step is determined as $\left\| \nabla_{\boldsymbol{w}_{\text{main}}} H_i \right\|$.
    
    \item These gradient norms are then sorted, and the unlabeled samples with the highest gradient norms are selected for annotation and inclusion in the labeled set \(\mathcal{L}_B^t\).

    \item These newly queried samples are annotated by the oracle and included in the labeled pool, which is subsequently used to train the main model.
\end{itemize}
The above steps will be repeated in each AL round $r$ until a termination criterion is met such as achieving a target accuracy.


\subsection{LPLgrad Training Phase}\label{sec:2}
The training phase of LPLgrad involves two primary models: the main model \(\boldsymbol{w}_{\text{main}}\) and the auxiliary model \(\boldsymbol{w}_{\text{aux}}\). LPLgrad trains the model \(\boldsymbol{w}_{\text{main}}\)  alongside   \(\boldsymbol{w}_{\text{aux}}\), which is integrated into its architecture to effectively capture intricate patterns and characteristics within input data. Here's how it works:
\begin{algorithm}[!t]
\caption{LPLgrad Training Phase}
\label{alg:algorithm}
\KwIn{Labeled pool $\mathcal{L}$, main model $\boldsymbol{w}_{\text{main}}$, auxiliary module $\boldsymbol{w}_{\text{aux}}$, number of AL rounds $\mathcal{T}$, number of epochs in each round $\mathcal{E}$}
\KwOut{$\boldsymbol{w}^T_{\text{main}}$}
Initialize $\boldsymbol{w}_{\text{main}}$,  $\boldsymbol{w}_{\text{aux}}$\\
\For{$t = 0, 1, 2, ..., \mathcal{T}$}{
    \For{$e = 0, 1, 2, ..., \mathcal{E}$}{
     Feedforward the input data of $\mathcal{L}$ to $\boldsymbol{w}_{\text{main}}$ and extract the features of input images as well as the output predictions of the $\boldsymbol{w}_{\text{main}}$.\\
     Calculate the loss of the main model  $l_{\text{main}}$ using equation~\eqref{eq:main}.\\
     Feedforward the extracted features to the $\boldsymbol{w}_{\text{aux}}$ to obtain $l_{\text{aux}}$ using equation~\eqref{eq:aux}.\\
     Add both of these losses $l_{\text{main}}$ and $l_{\text{aux}}$ using equation~\ref{eq:add}.\\
     Train both the $\boldsymbol{w}_{\text{main}}$ and $\boldsymbol{w}_{\text{aux}}$ in conjunction using the calculated losses.\\
    }
    Call Algorithm 2
}
\textbf{return} The final model $\boldsymbol{w}^T_{\text{main}}$ for the round $\mathcal{T}$ \\
\end{algorithm}

For each data point \( x_i \), we obtain two values: one is the prediction of the main model \( y_\text{main} = \boldsymbol{w}_\text{main}(x_i) \), and the other is a feature map \( F \), which is processed by auxiliary the model \(\boldsymbol{w}_{\text{aux}}\) to output the predicted loss \( l_\text{aux} = \boldsymbol{w}_\text{aux}(F) \). The loss of the main model \(\boldsymbol{w}_{\text{main}}\) is calculated using cross-entropy loss that takes the predicted value \( y_\text{main} \) and ground-truth label \( y_i \) of the sample \( x_i \) as inputs, which can be expressed as

\begin{equation}
\label{eq:main}
    l_\text{main} = \frac{1}{N} \sum_{i=1}^{N} \mathcal{L}_{\text{CE}}(y_i, y_\text{main})
\end{equation}

The loss for the auxiliary model \(\boldsymbol{w}_{\text{aux}}\) is computed based on the predicted loss \(l_\text{aux}\) and its corresponding ground-truth loss value \(l_\text{main}\), which can be presented as:

\begin{equation}
\label{eq:aux}
    l_\text{aux} = \frac{1}{P} \sum_{i=1}^{P} \max(0, M - \text{d}_i \cdot (l_{\text{aux}, i} - l_{\text{main}, i}))
\end{equation}

Here, \(M\) is a parameter for margin which explains how much the predicted loss should differ from the ground-truth loss before a penalty is applied. $d$ in the equation is used to determine the direction of the margin penalty and is computed as  
\begin{equation}
    \label{eq:direc}
    \text{d}_i = \text{max}(0, l_{\text{main}, i})
\end{equation}

 This ensures that the margin is adjusted correctly, either penalizing or not penalizing the predicted loss, depending on the relative difference between the predicted and true losses. 
To make the auxiliary model \(\boldsymbol{w}_{\text{aux}}\) robust to overall scale variations in the loss, we construct a mini-batch of \(P\) examples from \(\mathcal{L}_B\). We form \(P/2\) data pairs, denoted as \(\{ x^p = (x_m, x_n) \}\), where \(x^p\) represents a pair of examples \(m\) and \(n\). The superscript \(p\) indicates the loss for a pair of data points, denoted as \(l_{\text{aux}, i}\) and \(l_{\text{main}, i}\) for the auxiliary model and the main model, respectively, as shown in equation~\eqref{eq:aux}. 

Note that \(l_{\text{aux}, i}\) in equation~\eqref{eq:aux} represents the predicted loss for a specific sample in the pair, which is obtained by processing the extracted features of input through the \(\boldsymbol{w}_{\text{aux}}\) while the overall loss for auxiliary model is denoted by \(l_{\text{aux}}\). The auxiliary model \(\boldsymbol{w}_{\text{aux}}\) is learned by comparing the differences between the predicted losses \(l_\text{main}^p\) and \(l_\text{aux}^p\) for each data pair.

The total loss during the training phase is then calculated as follows:

\begin{equation}
\label{eq:add}
    L_\text{total} = l_\text{aux} + l_\text{main}
\end{equation}

LPLgrad leverages the extraction of multi-level input features obtained from various layers of the main model \(\boldsymbol{w}_{\text{main}}\), which are subsequently fed into the auxiliary model \(\boldsymbol{w}_{\text{aux}}\). Specifically, the model \(\boldsymbol{w}_{\text{aux}}\) comprises a series of blocks corresponding to the layers within the model \(\boldsymbol{w}_{\text{main}}\). Each block consists of two distinct layers: a global average pooling layer and a fully-connected layer. These\ blocks process the feature maps \( F \) derived from the layers of \(\boldsymbol{w}_{\text{main}}\) model, producing scalar values representing predicted losses \( l_\text{aux} \). Our objective is to jointly minimize both the predicted loss  \( l_\text{aux} \) generated by the model \(\boldsymbol{w}_{\text{aux}}\)  and the actual loss \( l_\text{main} \) determined by the model \(\boldsymbol{w}_{\text{main}}\). This optimization strategy enables the model to not only discern relevant input features but also effectively integrate rich, multi-level input space information.


Algorithm~\ref{alg:algorithm} summarizes the LPLgrad's training phase.

\subsection{LPLgrad Querying Phase}\label{sec:3}

\begin{algorithm}[tb]
\caption{ LPLgrad Querying Phase}
\label{alg:algorithm2}
 \KwIn{Unlabeled data pool $\mathcal{U}$, initial labeled pool $\mathcal{L}$, main model $\boldsymbol{w}_{\text{main}}$, number of AL rounds $\mathcal{T}$, annotation budget in each round $B$}
\KwOut{$\boldsymbol{w}^T_{\text{main}}$}

\For{t = 0, 1, 2, ....., ${T}$}{
     Calculate the entropy of each sample in $\mathcal{U}$ using equation ~\ref{eq:entropy}.\\
    Calculate the gradient of this calculated entropy with respect to main model $\boldsymbol{w}_{\text{main}}$ parameters using equation ~\ref{eq:grad} and subsequently its norm.\\
    Select $B$ samples with the highest values for gradient norm from $\mathcal{U}$.\\
    Annotate these selected data points and add them to $\mathcal{L}$.
}
Call algorithm 1\\
\textbf{return} main model parameters $\boldsymbol{w}^T_{\text{main}}$ for the round $\mathcal{T}$
\end{algorithm}

After training the main model $\boldsymbol{w}_{\text{main}}$  alongside the auxiliary model $\boldsymbol{w}_{\text{aux}}$, LPLgrad transitions to its second phase, which focuses on querying new samples for labeling. While the method proposed by Yoo et al. \cite{yoo2019learning} utilizes the $\boldsymbol{w}_{\text{aux}}$ model to identify the most informative points, our empirical results reveal a more effective strategy. Specifically, selecting samples based on their entropy values presents a robust alternative to the aforementioned approach. The hyperparameters in the loss prediction module can be highly sensitive in large-scale datasets like \cite{le2015tiny} and EMNIST \cite{cohen2017emnist} leading to performance degradation. Moreover, the model's uncertainty is better estimated using its entropy because it incorporates the model's confidence scores directly rather than with an attached loss prediction module. Specifically, selecting samples based on their gradient norm values presents a robust alternative to the aforementioned approach.

To implement this, we begin by extracting the embeddings of each sample in the unlabeled set \({x}_i \in \mathcal{U}_B\) such that \({z}_i = \boldsymbol{w}_\text{main}({x_i})\). We then use a softmax classifier to obtain the posterior probabilities \(P(y_i|{x}_i)\), which can be expressed as 

\begin{equation}
\label{softmax}
   P(y_i=c|{x}_i) = \frac{e^{z_c}}{\sum_{c=1}^{C} e^{z_{c}}}
\end{equation}
These posterior probabilities are then used to calculate the output entropy of each sample, which can be given as
\begin{equation}
\label{eq:entropy}
   H(P(y_i|{x}_i)) = - \sum_{c=1}^C P(y_i=c|{x}_i) \log P(y_i=c|{{x}_i})
\end{equation}
where \(P(y_i=c|{x}_i)\) is the predicted probability for class \(c\) given the sample \({x}_i\), and \(C\) is the total number of classes in the dataset.

We treat this entropy as a loss and perform a backward pass on the loss function to compute the gradient of the $\boldsymbol{w}_{\text{main}}$ model parameters for each sample \({x}_i\):

\begin{equation}
\label{eq:grad}
   \nabla_{\boldsymbol{w}_{\text{main}}} H_i = \frac{\partial H(P(y_i|{x}_i))}{\partial \boldsymbol{w}_{\text{main}}} \bigg |_{{x}_i}
\end{equation}
The Frobenius norm of these gradients is then calculated to assess the network's sensitivity to the input as 

\begin{equation}
\label{eq:norm}
   g_i = \left\| \nabla_{\boldsymbol{w}_{\text{main}}} H_i \right\|_{F}
\end{equation}
These gradient norms of all the inputs are subsequently stored and sorted to identify the highest values, which can be represented as

\begin{equation}
\label{eq:sort}
   \{ g_{i_1}, g_{i_2}, \ldots, g_{i_B} \} \quad \text{s.t.} \quad g_{i_1} \geq g_{i_2} \geq \ldots \geq g_{i_B}
\end{equation}
In our querying phase, we follow the theoretical insights presented in \cite{wang2022boosting}, which suggest that selecting samples with higher gradient norms from the unlabeled set can lead to a reduction in the upper bound of the total loss. Thus, we guide our selection process by prioritizing samples with the largest gradient norms for annotation. Subsequently, these selected samples are added to the labeled set \(\mathcal{L}\). Importantly, the parameters of the ${\boldsymbol{w}_{\text{main}}}$  model will not be updated during the selection process of the new samples.

Algorithm~\ref{alg:algorithm2} summarizes the LPLgrad's querying phase.

\section{Experiments}\label{sec:Exp}
In this section, we evaluate the performance of LPLgrad across various datasets and compare its accuracy with several baseline methods. Additionally, we assess LPLgrad's computational performance, including both querying and training time.

\subsection{Experimental Setup}
To ensure accurate results, we average all the experiments over 5 trials and report the mean outcomes. Below, we describe our setup in detail:

{\bf Datasets.} We evaluate our proposed approach, LPLgrad, using four publicly available benchmark datasets: CIFAR-10 \cite{CIFAR-10}, CIFAR-100 \cite{CIFAR-10}, SVHN \cite{netzer2011reading}, Caltech-101 \cite{FeiFei2004LearningGV}, which are commonly used in state-of-the-art (SOTA) comparisons. In addition, We validate LPLgrad on another real-world dataset, {\em comprehensive disaster dataset (CDD)} \cite{verma2021disaster}, to assess its robustness. Below, we provide a brief description of each dataset.
\begin{itemize}
    \item \textbf{CIFAR-10 \& CIFAR-100 \cite{CIFAR-10}:} are two datasets, with the former containing 10 different classes and the latter containing 100 classes. Each dataset comprises 60,000 color images, evenly distributed across the classes. Specifically, in CIFAR-10, each class has 5000 samples, whereas in CIFAR-100, each class has 600 samples. The images are all 32$\times$32 pixels in size and feature various objectives such as animals, vehicles, etc. 

    \item \textbf{SVHN \cite{netzer2011reading}:} The SVHN (Street View House Numbers) dataset comprises a total of 630,420 colored images of house numbers, categorized into 10 classes for each digit. The images are of size 32$\times$32, similar to CIFAR-10 and CIFAR-100. The dataset is divided into three sets: 73,257 images for training, 26,032 images for testing, and an extra set containing 530,420 images. For a fair comparison with other AL methods, we do not use the extra set provided in the SVHN dataset.

   \item {\bf Caltech-101 \cite{FeiFei2004LearningGV}:} It contains images of objects belonging to 101 categories, making it a highly imbalanced dataset with over 100 classes. For instance, there are about 40 to 800 images per category, with most categories having around 50 images. It contains 5800 training while 2877 test images with each image being roughly 300 $\times$ 200 pixels in size.

   \item  {\bf Comprehensive Disaster Dataset (CDD) \cite{verma2021disaster}:} This dataset consists of a total of 10,733 images, categorized into fire disaster, human damage, land disaster, water disaster, and etc. It is divided into two main sets: 8,591 training images and 2,142 test images, with a total of 6 classes. It is an extremely imbalanced dataset, with the number of samples per class varying from 29 to 1,668. These images have a resolution of 32 $\times$ 32 pixels.
\end{itemize}

\begin{table}[!t]
\setlength{\tabcolsep}{0.6em}
\centering
\renewcommand{\arraystretch}{1.2}
\caption{Datasets, Models, and Training Parameters.} 
\label{tab:parameters}

\resizebox{\linewidth}{!}{%
 \begin{tabular}{p{1.33cm}|p{0.7cm}|p{0.8cm}|p{1.22cm} |p{1.2cm}| p{0.7cm}|p{1.23cm}}
 \bottomrule 
Dataset &Size&Classes&Model&Parameters& Epochs& mini-batch\\
 \hline 
CIFAR 10 &70,000&10&ResNet-18&1,100,000&200&128\\
 \hline 
{\footnotesize CIFAR-100}&70,000&10&ResNet-18&1,100,000&200&128\\
 \hline
  SVHN &73257&10&ResNet-18&1,100,000&200&128\\
\hline
 Caltech101 &6110&101&ResNet-18&1,100,000&50&80\\  
 \hline
 CDD   &8677&6&ResNet-18&1,100,000&50&128\\
 \hline\hline
 \multicolumn{3}{c||}{$\eta=0.001-0.1$}&\multicolumn{4}{c}{SGD Momentum$=0.9$}\\
 \toprule
\end{tabular}} 
\end{table}

{\bf Baselines.} We compare our proposed LPLgrad algorithm with state-of-the-art AL approaches discussed in the literature \cite{luo2013latent, sener2017active, wang2022boosting, yoo2019learning}. Specifically, we consider the two main categories: i) uncertainty-based methods such as LearningLoss \cite{yoo2019learning}, Ent-GradNorm \cite{wang2022boosting}, and entropy \cite{wang2014new}, and ii) diversity-based methods such as coreset \cite{sener2017active}, which have been reviewed in Sections~\ref{sec:intro} and \ref{sec:related}. In addition, we compare LPLgrad with non-AL strategies, where unlabeled samples are selected randomly to augment the labeled set. We refer to this method as random sampling (Rand). 

\begin{table}[!t]
    \centering
    \caption{Values of \( B \), \( R \) and $\mathsf{A}$ for all datasets.}
    \label{tab:br_values}
    \begin{tabular}{cccccc}
        \toprule
        & \textbf{CIFAR10} & \textbf{CIFAR100} & \textbf{SVHN} & \textbf{Caltech101} & \textbf{CDD} \\
        \midrule
        \textbf{$\mathbf{B}$} & 1,000 & 2,500 & 1,000 & 500 & 500 \\
        \textbf{$\mathbf{R}$} & 25,000 & 25,000 & 25,000 & 2,000 & 1,000 \\
       $\mathsf{A}$ & 10,000 & 20,000 & 10,000 & 3,500 & 2,500 \\ 
        \midrule
        \textbf{$\mathbf{B}$} & 500 & 1,000 & 500 & 200 & 200 \\
        \textbf{$\mathbf{R}$} & 5,000 & 15,000 & 10,000 & 1,000 & 1,000 \\
        $\mathsf{A}$ & 5,000 & 8,000 & 5,000 & 1,400 & 1,000 \\ 
        \bottomrule
    \end{tabular}
\end{table}

\begin{table*}[!t]
\centering
\setlength{\tabcolsep}{1em}
\renewcommand{\arraystretch}{1.2}
\caption{Comparison of LPLgrad vs. baselines on all datasets: CIFAR10, CIFAR100, SVHN, and Caltech-101 for high $\mathsf{A}$.}
\label{tab:res}
\resizebox{\linewidth}{!}{
\begin{tabular}{l|c|c|c|c|c|c|c|c|c|c|c|c }
\toprule
& \multicolumn{3}{c|}{\textbf{CIFAR10}} &  \multicolumn{3}{c|}{\textbf{CIFAR100}} &  \multicolumn{3}{c|}{\textbf{SVHN}} & 
\multicolumn{3}{c}{\textbf{Caltech-101}} \\ 
\cmidrule{2-4} \cmidrule{5-7} \cmidrule{8-10} \cmidrule{11-13}
\textbf{Actively chosen data} & \textbf{3000} & \textbf{6000} & \textbf{9000} & \textbf{7500} & \textbf{12500} & \textbf{20000} &  \textbf{3000} & \textbf{6000} & \textbf{9000}  & \textbf{1500} & \textbf{2500} & \textbf{3500} \\ 
\midrule
LearningLoss & 71.3 $\pm$ 1.1 & 86.0 $\pm$ 0.2 & 89.7 $\pm$ 0.3 & 43.7 $\pm$ 1.1 & 60.6 $\pm$ 0.4 & 66 $\pm$ 0.5  & 90 $\pm$ 0.3 & {93.9 $\pm$ 0.1} & 95.1 $\pm$ 0.1  & 36.0 $\pm$ 0.6 & 46.8 $\pm$ 0.5 & 53.5 $\pm$ 1.0 \\
Entropy & 65.2 $\pm$ 2.9 & 84.1 $\pm$ 0.8 & 88.9 $\pm$ 0.4 &  44.4 $\pm$ 1.0 & 60.4 $\pm$ 0.3 & 66.5 $\pm$ 0.2 & 89.6 $\pm$ 0.3& 93.7 $\pm$ 0.1 & 94.9 $\pm$ 0.1 & 36.9 $\pm$ 0.4 & 45.6 $\pm$ 0.8 & 54.1 $\pm$ 1.5\\

Ent-GradNorm & 68.0 $\pm$ 1.3 & 84.8 $\pm$ 0.4 & 89.0 $\pm$ 0.2 & 46.6 $\pm$ 0.8 & 61.7 $\pm$ 0.5 & 66.7 $\pm$ 0.2 & 89.7 $ \pm$ 0.3  &93.4 $ \pm$ 0.1  & 94.8 $ \pm$ 0.1 & 38.3 $\pm$ 1.4 & 47.5 $\pm$ 1.6 & 54.4 $\pm$ 1.4 \\

Coreset & 64.0 $\pm$ 1.9 & 81.7 $\pm$ 0.6 & 86.0 $\pm$ 0.3 &  44.3 $\pm$ 1.0 & 59.8 $\pm$ 0.8 & 64.5 $\pm$ 0.6 & 82.3 $\pm$ 0.3 &87.6 $\pm$ 0.1  & 89.7 $\pm$ 0.1 & 37.8 $\pm$ 1.3 & 45.5 $\pm$ 2.1 & 51.5 $\pm$ 2.0 \\

Rand & 71.7 $\pm$ 0.8 & 83.8 $\pm$ 0.7 & 86.2 $\pm$ 0.5 &  44.4 $\pm$ 1.8 & 59.3 $\pm$ 0.6 & 64.3 $\pm$ 0.1 & 81.9 $\pm$ 0.5& 87.5 $\pm$ 0.1 &89.6 $\pm$ 0.0 & 36.0 $\pm$ 1.4 & 44.1 $\pm$ 1.6 & 50.8 $\pm$ 1.7 \\
\midrule
\rowcolor{red!10} {\bf LPLgrad (Ours)} & {72.4 $\pm$ 1.1} & {86.2 $\pm$ 0.2} & {90.1 $\pm$ 0.3} &{46.8 $\pm$ 1.1} & {62.8 $\pm$ 0.4} & {67 $\pm$ 1.4} & {90.2 $\pm$0.1} & {93.7$\pm$0.}1& {95.1$\pm$0.0}& {38.6 $\pm$ 1.9} & {49.1 $\pm$ 2.0} & {56.6 $\pm$ 1.2}  \\
\bottomrule
\end{tabular}}
\end{table*}
{\bf Training Models. } To ensure a fair comparison with our baselines, we train an 18-layer residual network (ResNet-18) as our main model $\boldsymbol{w}_{main}$ to perform image classification across all the experiments. We adhere to the settings and hyperparameters recommended by the baseline studies for reproducing their results. Additionally, we employ stochastic gradient descent (SGD) as the optimizer, with a momentum of 0.9 for all datasets. The rest of the training settings for each dataset are as follows:

\begin{itemize}
    \item For CIFAR10 and CIFAR100, we use a learning rate (lr) of 0.1 and train both $\boldsymbol{w}_{main}$ and $\boldsymbol{w}_{aux}$ models for 200 epochs, reducing the lr by a factor of 0.1 after 160 epochs.

    \item For SVHN, we use a lr of 0.01 and train both $\boldsymbol{w}_{main}$ and $\boldsymbol{w}_{aux}$ models for 200 epochs, reducing the lr by 0.1 after 40 epochs.

    \item For Caltech-101, the $\boldsymbol{w}_{main}$ and $\boldsymbol{w}_{aux}$ models are trained for 50 epochs with a lr of 0.01, which is decayed by 0.1 after 40 epochs.

    \item For CDD, the $\boldsymbol{w}_{main}$ and $\boldsymbol{w}_{aux}$ models are trained for 200 epochs with a lr of 0.01, which is decayed by 0.1 after 40 epochs.
\end{itemize}  
For training the ResNet-18 model, we begin with a randomly selected initial pool of samples $B$ from the  large unlabeled pool to be annotated and transferred to the labeled set, which is then added to the training dataset. However, we note that this random selection might result in overlapping images and similar selections each time. To mitigate this, we employ the technique used in \cite{beluch2018power} to obtain a random subset $\mathcal{S}_\text{R} \subset \mathcal{U}_\text{N}$ from the pool of unlabeled samples. This simple technique proves to be efficient as it reduces redundancy. The value of $R$ is adjusted for each dataset, depending on the total number of samples and the size of $B$, considering differences in dataset size and class imbalance, as each class within the dataset has a different number of data points.

\begin{figure}[!t]
  \centering
  \begin{minipage}[t]{0.5\linewidth}
    \centering
    \begin{adjustbox}{valign=t}
      \includegraphics[width=\linewidth]{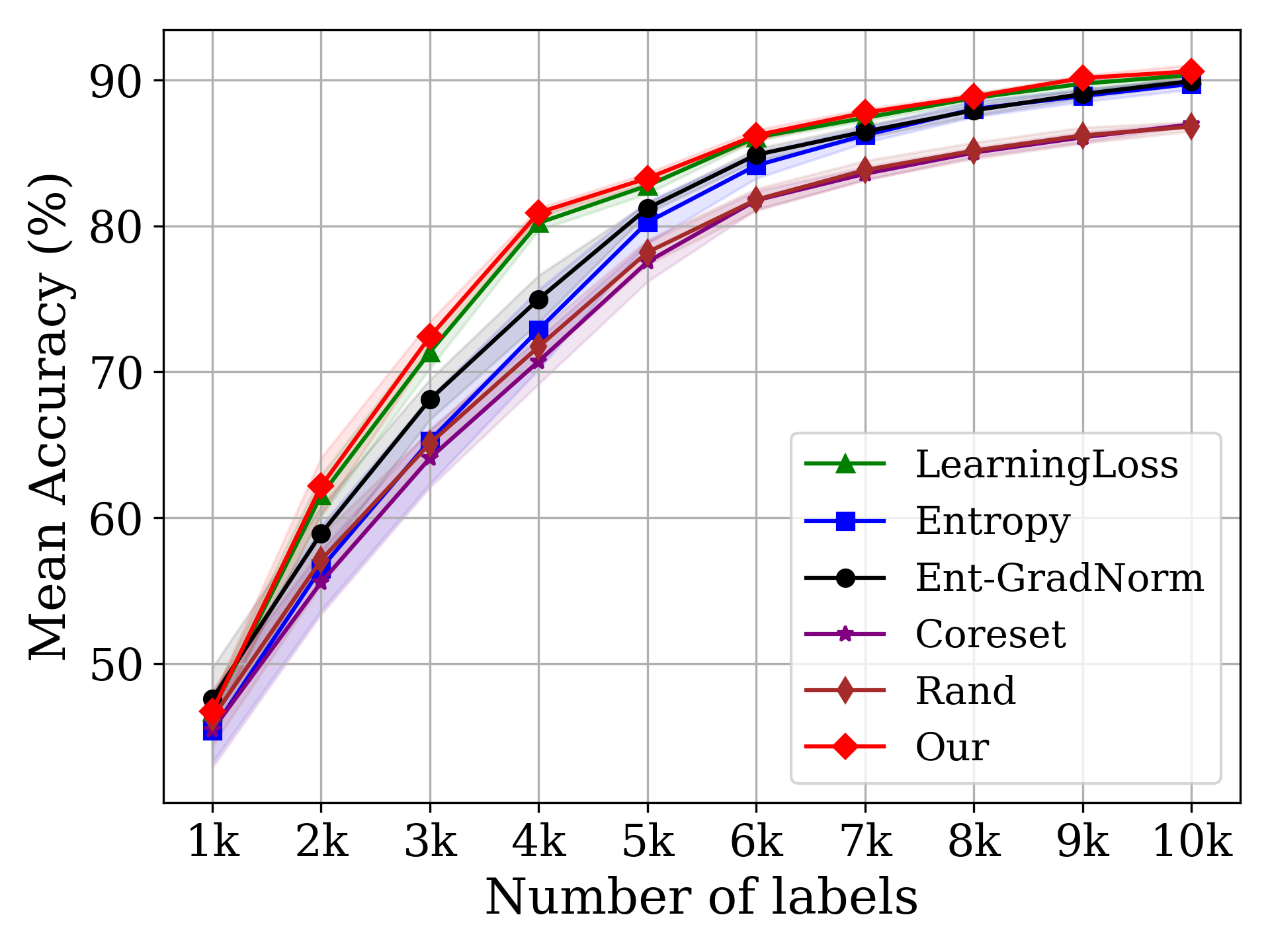}
    \end{adjustbox}
    \caption*{(a) CIFAR-10 dataset.}
    \label{fig:CIF10_1}
  \end{minipage}\hfill
  \begin{minipage}[t]{0.5\linewidth}
    \centering
    \begin{adjustbox}{valign=t}
      \includegraphics[width=\linewidth]{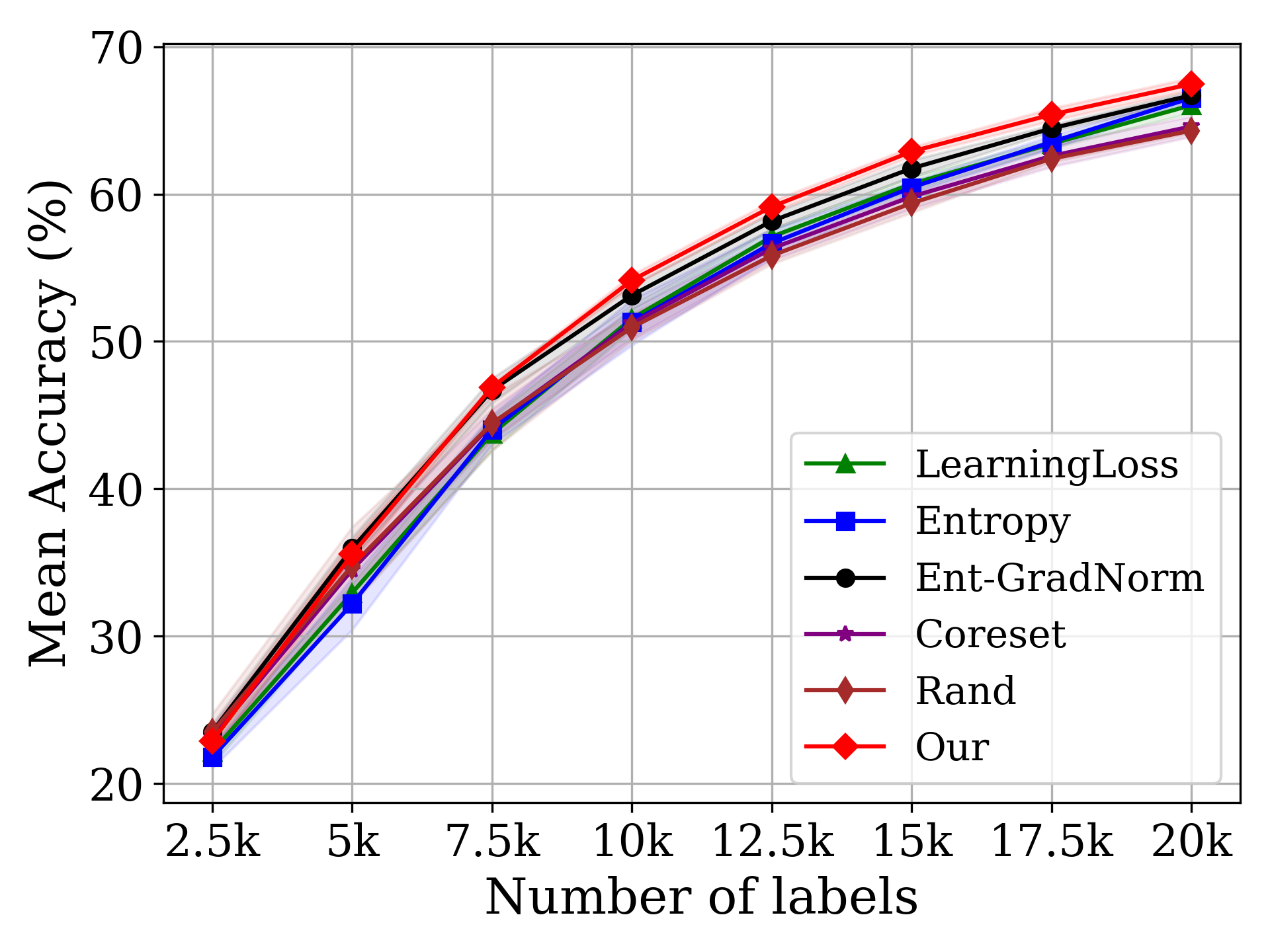}
    \end{adjustbox}
    \caption*{(b) CIFAR-100 dataset.}
    \label{fig:CIF100_1}
  \end{minipage}\\[2ex]
  \begin{minipage}[t]{0.5\linewidth}
    \centering
    \begin{adjustbox}{valign=t}
      \includegraphics[width=\linewidth]{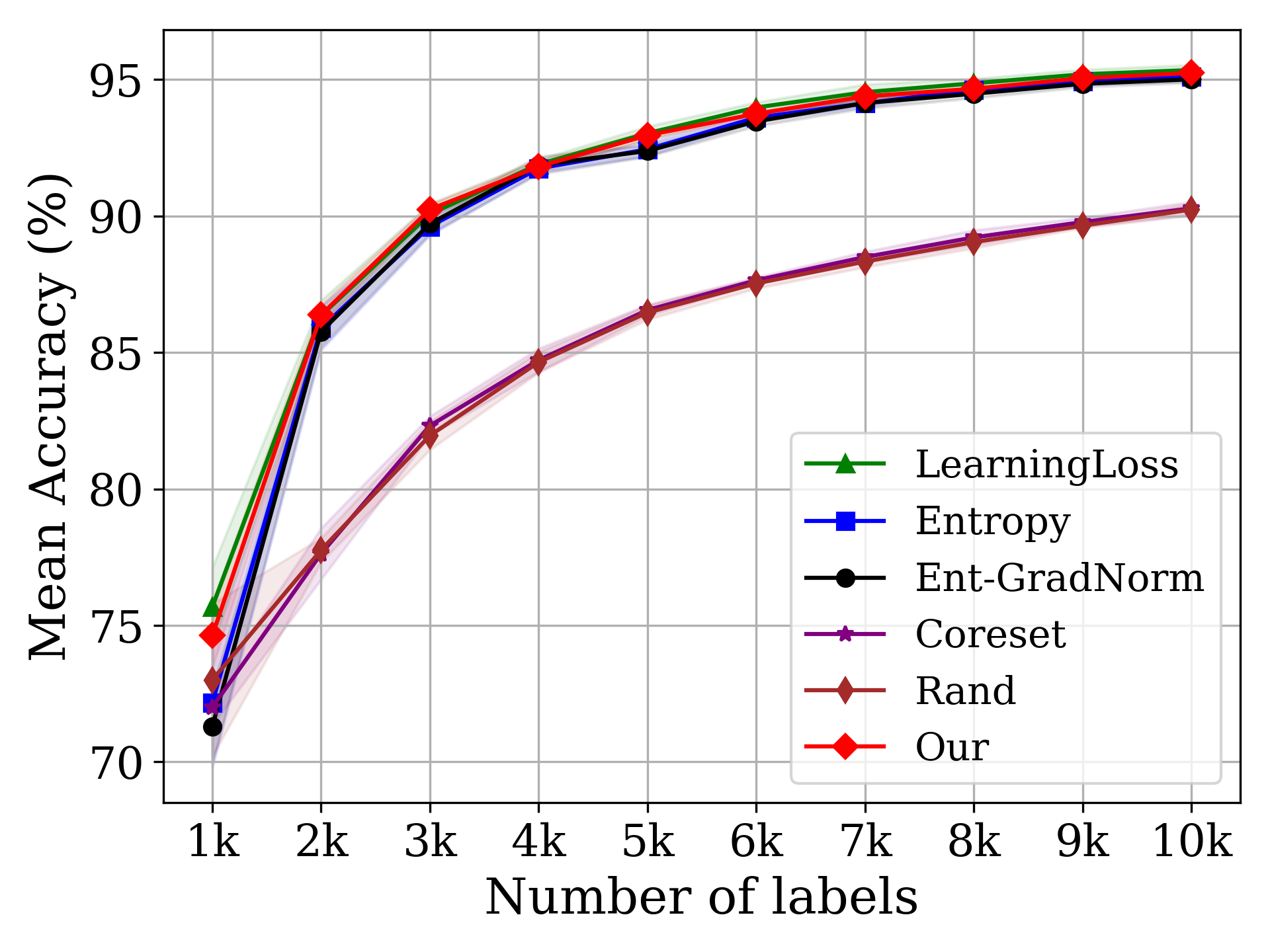}
    \end{adjustbox}
    \caption*{(c) SVHN dataset.}
    \label{fig:SVHN_1}
  \end{minipage}\hfill
  \begin{minipage}[t]{0.5\linewidth}
    \centering
    \begin{adjustbox}{valign=t}
      \includegraphics[width=\linewidth]{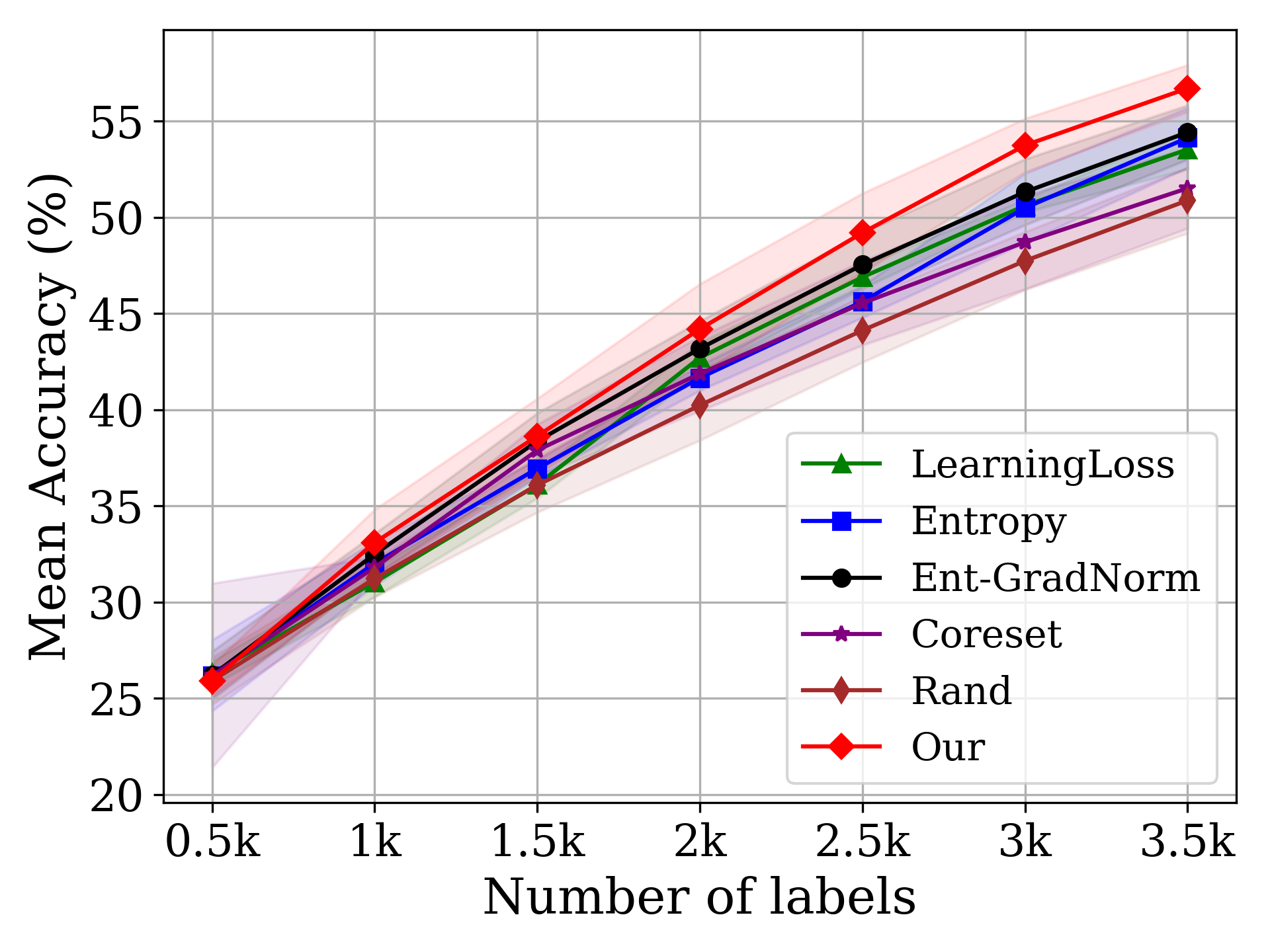}
    \end{adjustbox}
    \caption*{(d) Caltech-101 dataset.}
    \label{fig:Cal101_1}
  \end{minipage}
  \caption{Classification performance of LPLgrad compared to baseline methods on CIFAR-10, CIFAR-100, SVHN, and Caltech-101 datasets with high \(\mathsf{A}\).}
  \label{fig:Result}
\end{figure}

In Table~\ref{tab:br_values}, we detail the query batch size \(B\), the size of the randomly chosen set \(R\), and the annotation budget ($\mathsf{A}$) for all experiments across all datasets.  The values for $\mathsf{A}$ and \(R\) have been chosen following the standard practice in AL literature \cite{kothawade2021similar, hacohen2022active, zhao2021active, wan2021nearest}. The upper part of the table provides details for a high $\mathsf{A}$ (with plots shown in \fref{fig:Result}), while the lower part corresponds to a low $\mathsf{A}$ (with plots shown in \fref{fig:Result_2}).

\subsection{Results}

{\bf Comparing LPLgrad with baselines.} We first evaluate the performance of our LPLgrad approach against the baselines in terms of image classification with a high $\mathsf{A}$. In Table~\ref{tab:res}, we present detailed results showing that LPLgrad consistently outperforms its counterparts across almost all datasets. Notably, LPLgrad achieves a significant margin of superiority, especially with challenging datasets such as CIFAR-100 and Caltech101. For example, on the Caltech101 dataset, which is highly imbalanced with over 100 classes, LPLgrad demonstrates a clear advantage over all baselines, improving accuracy by approximately 5\%. LearningLoss \cite{yoo2019learning} emerges as the second-best performing method, sometimes surpassing Ent-GradNorm \cite{wang2022boosting}. Entropy often ranks third, followed by Coreset and Rand. Interestingly, Coreset performs similarly to Rand in almost all scenarios.

\begin{figure}[!t]
  \centering
  \begin{minipage}[!t]{0.5\linewidth}
    \centering
    \begin{adjustbox}{valign=t}
      \includegraphics[width=\linewidth]{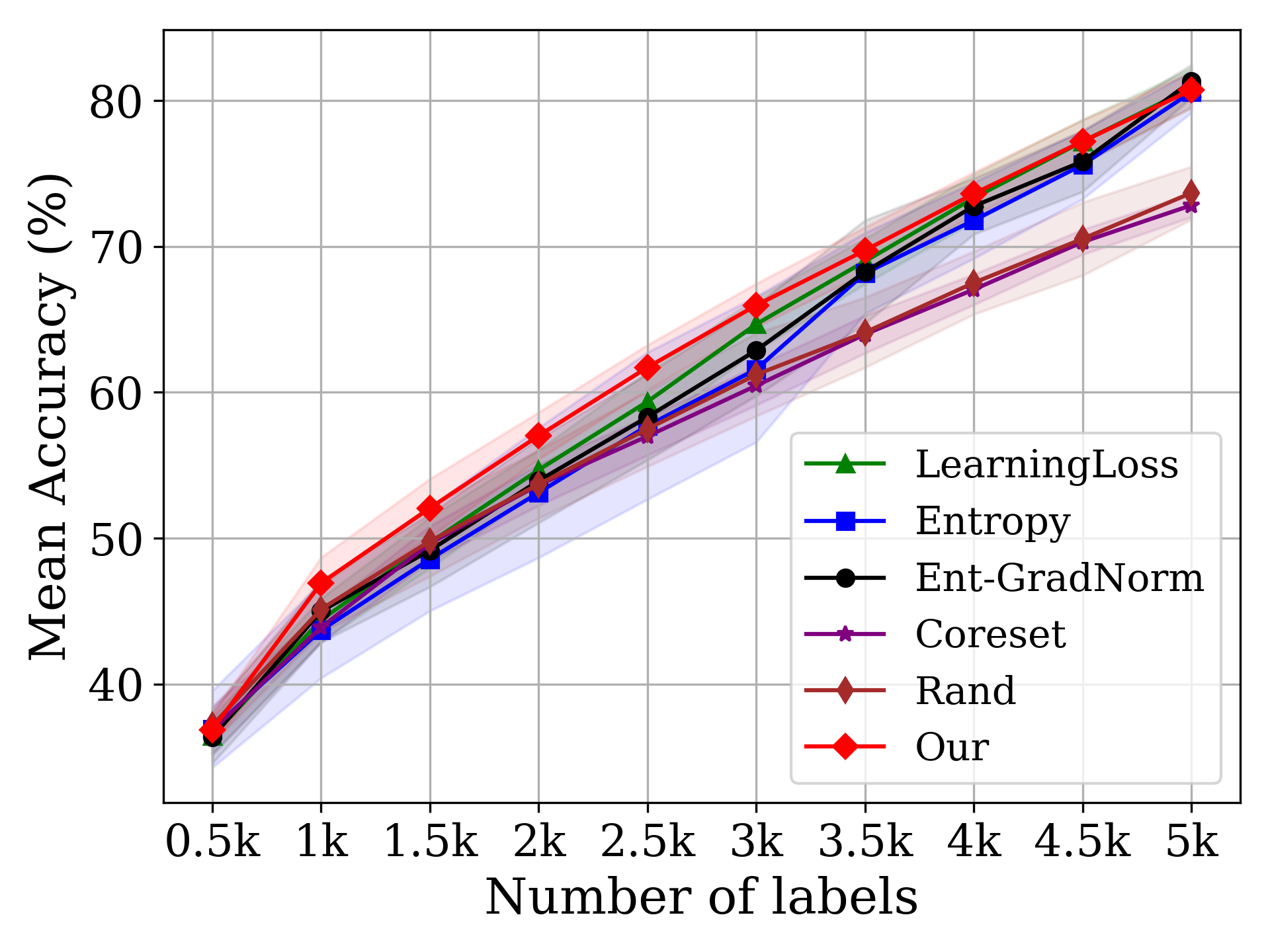}
    \end{adjustbox}
    \caption*{(a) CIFAR-10 dataset.}
    \label{fig:CIF10}
  \end{minipage}\hfill
  \begin{minipage}[!t]{0.5\linewidth}
    \centering
    \begin{adjustbox}{valign=t}
      \includegraphics[width=\linewidth]{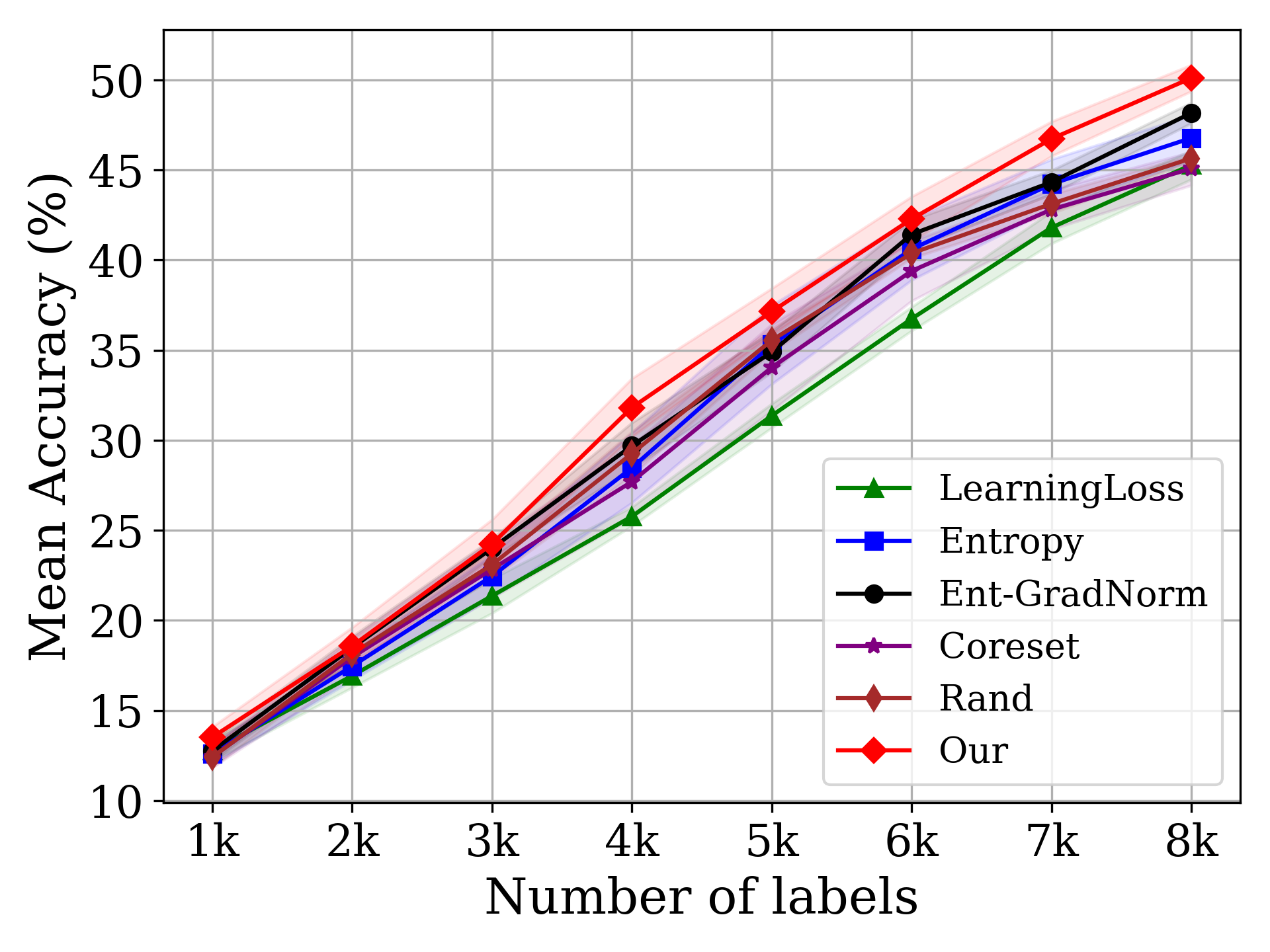}
    \end{adjustbox}
    \caption*{(b) CIFAR-100 dataset.}
    \label{fig:CIF100}
  \end{minipage}\\[2ex]
  \begin{minipage}[!t]{0.5\linewidth}
    \centering
    \begin{adjustbox}{valign=t}
      \includegraphics[width=\linewidth]{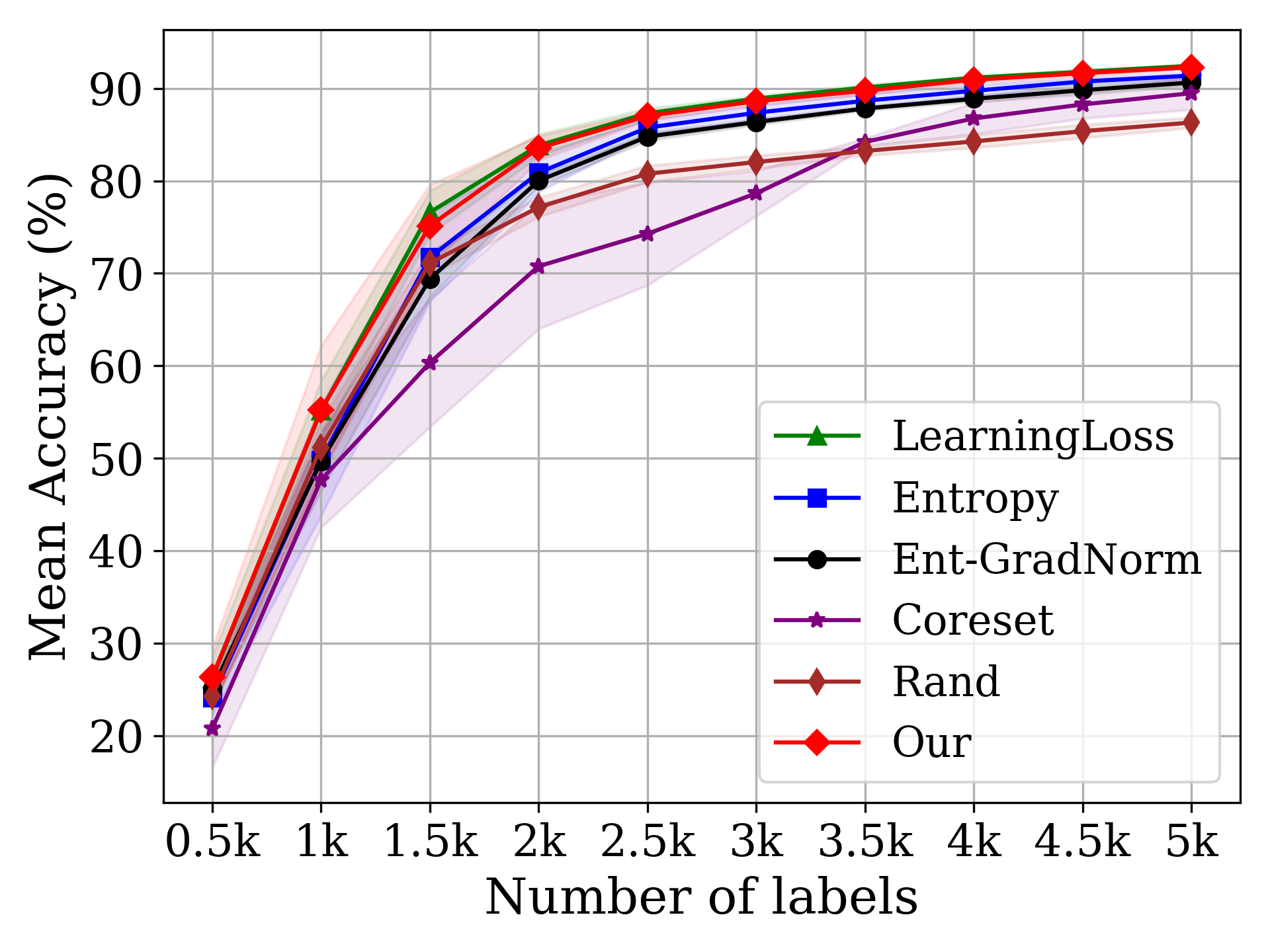}
    \end{adjustbox}
    \caption*{(c) SVHN dataset.}
    \label{fig:SVHN}
  \end{minipage}\hfill
  \begin{minipage}[!t]{0.5\linewidth}
    \centering
    \begin{adjustbox}{valign=t}
      \includegraphics[width=\linewidth]{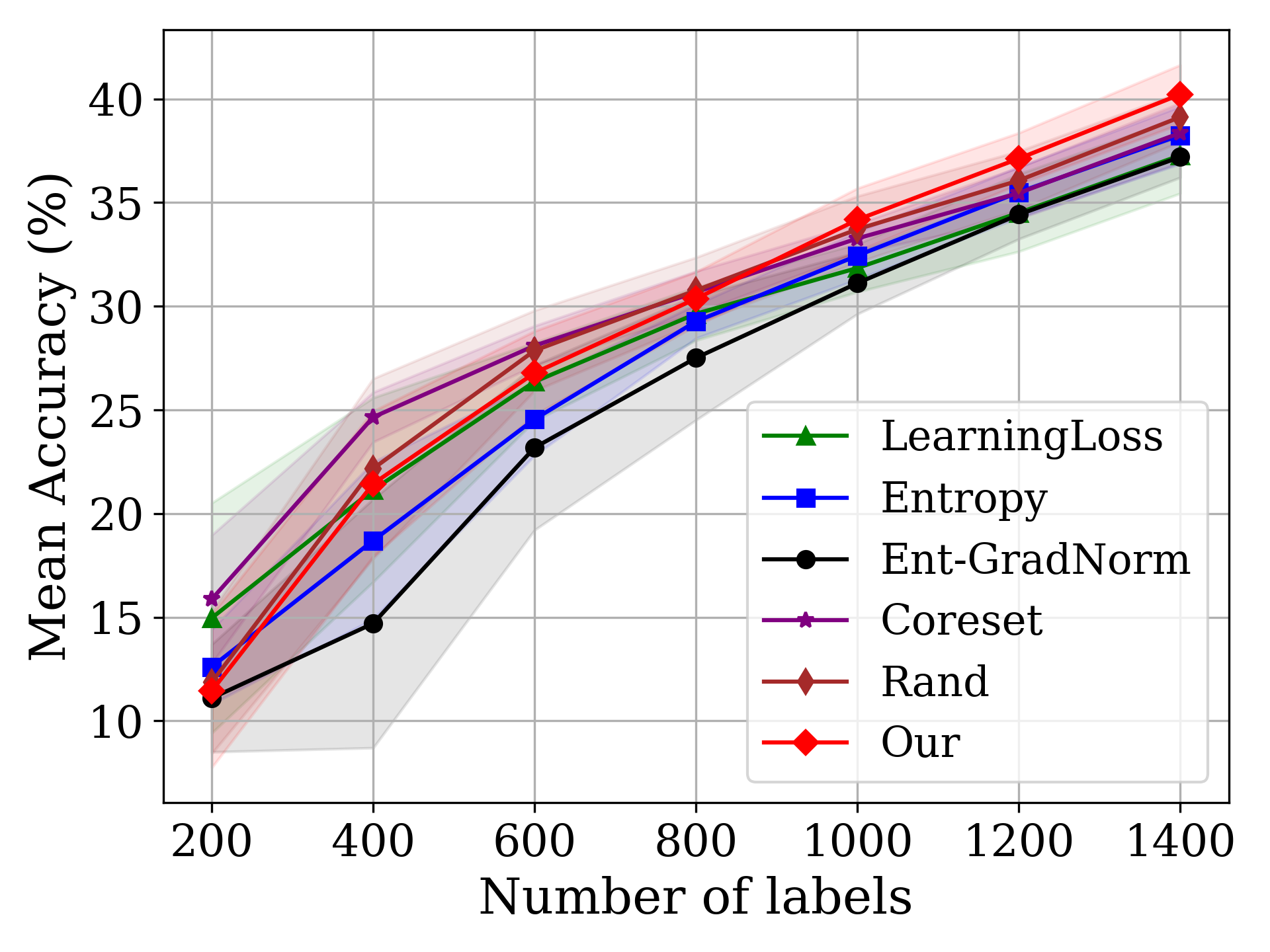}
    \end{adjustbox}
    \caption*{(d) Caltech-101 dataset.}
    \label{fig:Cal101}
  \end{minipage}
  \caption{Classification performance of LPLgrad compared to baseline methods on CIFAR-10, CIFAR-100, SVHN, and Caltech-101 datasets with low \(\mathsf{A}\). }
  \label{fig:Result_2}
\end{figure}

\begin{table*}[!t]
\caption{Comparison of computational overheads (in seconds) of our LPLgrad against baselines across different datasets.}

\label{table:performance_comparison}
 
\setlength{\tabcolsep}{1em}
\renewcommand{\arraystretch}{1.2}
\centering
\begin{adjustbox}{max width=\textwidth}
\begin{tabular}{c|cc|cc|cc|cc|cc}
\toprule
\multirow{2}{*}{} & \multicolumn{2}{c|}{{\bf CIFAR10}} & \multicolumn{2}{c|}{{\bf CIFAR100}} & \multicolumn{2}{c|}{{\bf SVHN}} & \multicolumn{2}{c|}{{\bf Caltech101}} & \multicolumn{2}{c}{{\bf CDD}} \\ \cline{2-11}
 & Train Time & Querying Time & Train Time & Querying Time & Train Time & Querying Time & Train Time & Querying Time & Train Time & Querying Time \\ \hline
LearningLoss & 63.7 & 15.7  & 99.7 &95  &124.9  &210  & 280.9 & 75 &63.9  &40  \\ \hline
Entropy &61.9  &14.29  &97.5  &93  &124.4  &{\bf \underline{200.4}}  &252.5  &50  & 59.0 &35  \\ \hline
Ent-GradNorm &  63.1& 40 & 97.9 &115  & 135 &202.4  & {\bf \underline{182.5}} &56.6  & {\bf \underline{55.7}} &37.4  \\ \hline
Coreset & 68.42 &2319  &97.1  &1123  &130 & 2124 &197.4   & 676 & 63.1 &54.5  %
\\  \hline
\rowcolor{red!10} {\bf LPLgrad (Ours)} &  {\bf \underline{60.5}}& {\bf \bf \underline{14} }& {\bf \underline{92}} & {\bf\underline{92.3}} & {\bf \underline{123}} &215  &{252}  & {\bf\underline{49.1}}  &60  &{\bf \underline{34.8}}  \\ 
\bottomrule
\end{tabular}
\end{adjustbox}
\end{table*}

{\bf Evaluating the effect of $\mathsf{A}$.} In \fref{fig:Result} and \ref{fig:Result_2}, we present the classification accuracy of LPLgrad across various datasets under two different annotation budget (\(\mathsf{A}\)) settings: high and low. These results are compared against our baseline approaches. For instance, in \fref{fig:Result}(a), we show the accuracy curves of our baselines when classifying the CIFAR-10 dataset. Our approach, LPLgrad, outperforms the baselines across all \(\mathsf{A}\)s, achieving the highest accuracy of 91\% by the final round. LearningLoss~\cite{yoo2019learning} is the second-best performing method, followed by Ent-GradNorm~\cite{wang2022boosting} and entropy~\cite{wang2014new}. Entropy lags behind Ent-GradNorm for the first seven rounds but catches up in the last three rounds. Coreset~\cite{sener2017active}, the only diversity-based method, performs worse than all other methods in the initial half of the AL rounds, eventually aligning with Rand in the latter half.

For the CIFAR-100 dataset, which is the most challenging due to its 100 fine-grained classes, \fref{fig:Result}(b) demonstrates that LPLgrad has superior performance compared to all baselines, reaching a high accuracy of 68\%. Here, Ent-GradNorm is the second-best performing algorithm, followed by entropy sampling, indicating their effectiveness in a large-scale dataset setting. LearningLoss performs comparably to entropy until the end. Coreset and Rand are the worst-performing methods. As a purely diversity-based method, Coreset does not outperform the uncertainty-based methods like entropy, Ent-GradNorm, and LPLgrad.

\begin{figure}[!t]
  \centering
  \begin{minipage}[t]{0.24\textwidth}
    \centering
    \includegraphics[width=\textwidth]{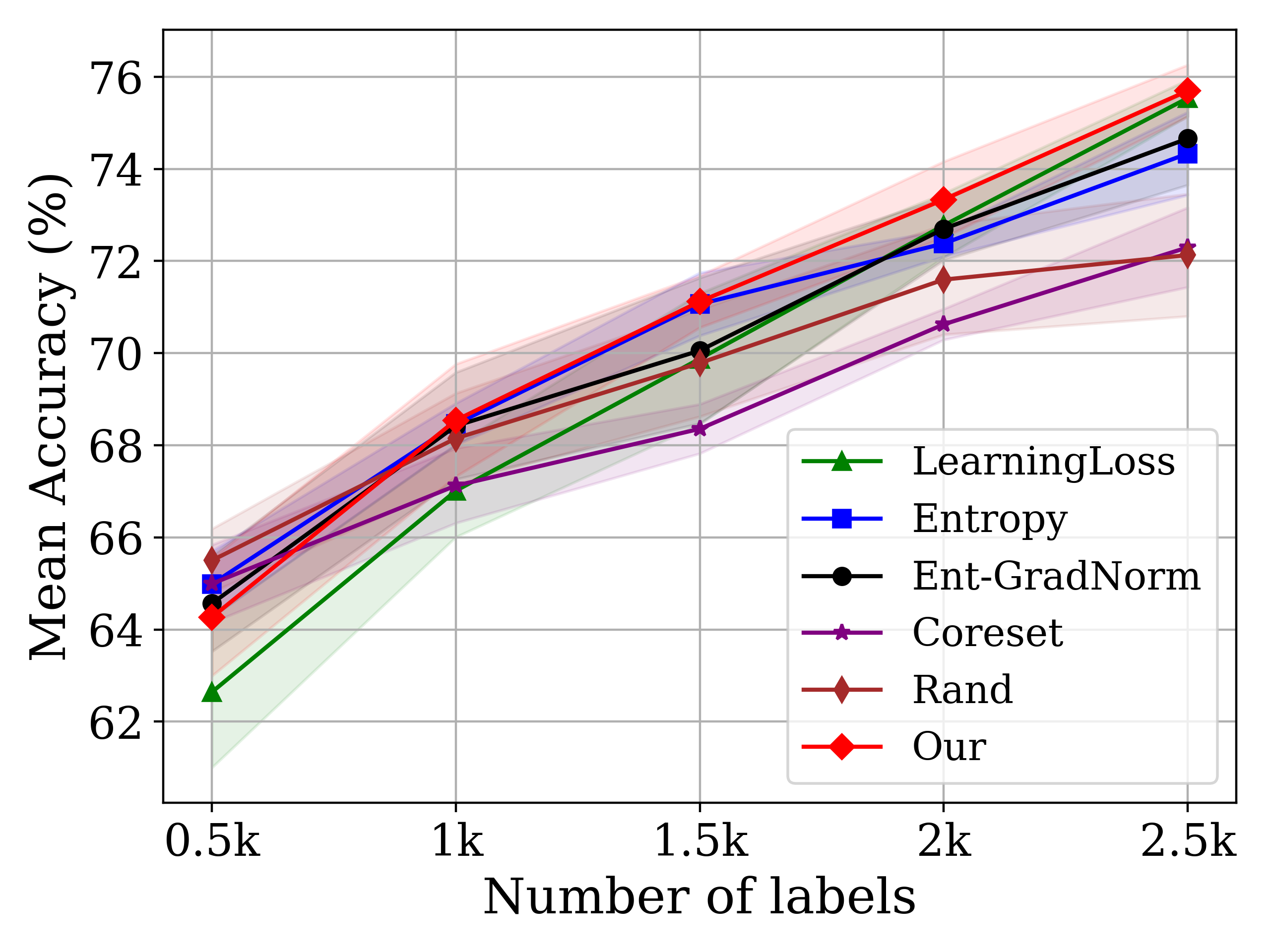}
  \end{minipage}\hfill
  \begin{minipage}[t]{0.24\textwidth}
    \centering
    \includegraphics[width=\textwidth]{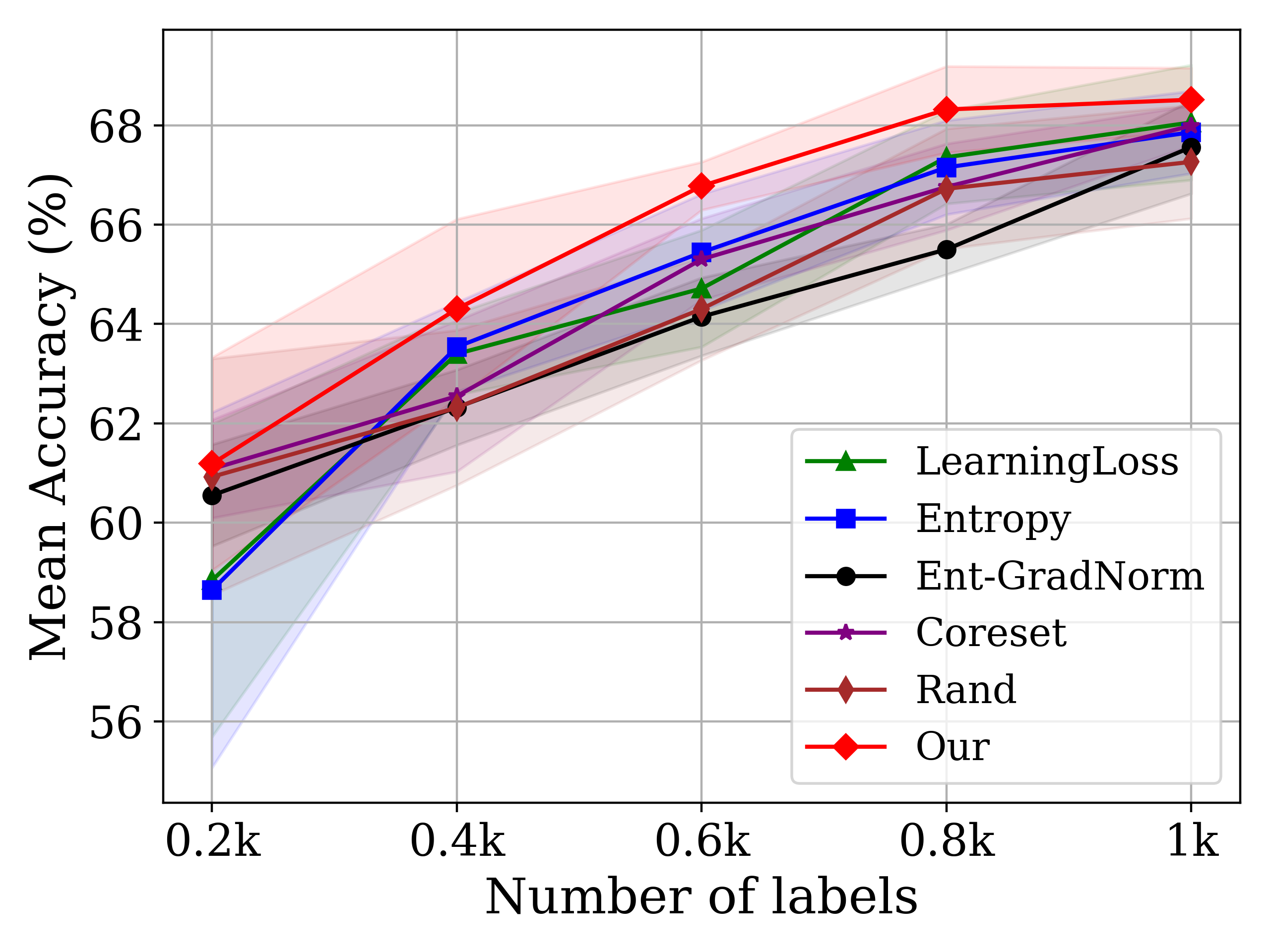}
  \end{minipage}\\[2ex]
    
  \caption{Classification performance comparison on Comprehensive Disaster Dataset.}
  \label{fig:disas}
\end{figure}

\fref{fig:Result}(c) represents the accuracy curves of LPLgrad on the SVHN dataset. All algorithms perform well on this dataset, including our LPLgrad approach, achieving \(\geq 95\%\) accuracy, except for Coreset and Rand, which achieve an accuracy of 90\%. LearningLoss performs worse than LPLgrad in the first half of the rounds but slightly leads in the later rounds. Entropy and Ent-GradNorm dip around the fifth AL round but stabilize and perform on par with other algorithms. Finally, \fref{fig:Result}(d) compares LPLgrad against the baselines on the Caltech-101 dataset. LPLgrad clearly outperforms the others, achieving 57.4\% accuracy. Ent-GradNorm follows as the second-best performer. Entropy and LearningLoss initially perform poorly but improve and end up close to Ent-GradNorm. Coreset is the second-worst AL method, followed by Rand. Uncertainty-based methods consistently lead over diversity-based methods. However, all approaches, including LPLgrad, show the lowest performance on the Caltech-101 dataset compared to the other datasets, suggesting that increasing the \(\mathsf{A}\) could improve results, albeit with more training time needed.

Examining the results with a low $\mathsf{A}$ in \fref{fig:Result_2}, LPLgrad consistently achieves the highest accuracy, outperforming all baselines across all datasets. An exception is observed in the Caltech-101 dataset, where the Coreset method initially achieves higher accuracy than LPLgrad. However, as the learning curves converge, LPLgrad again surpasses the baseline methods in accuracy. 

\begin{figure}[!t]
  \centering
  \begin{minipage}[t]{0.23\textwidth}
    \centering
    \includegraphics[width=\textwidth]{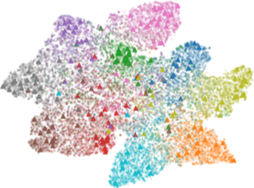}

    \caption*{(a) Round 2}
  \end{minipage}\hfill
  \begin{minipage}[t]{0.23\textwidth}
    \centering
    \includegraphics[width=\textwidth]{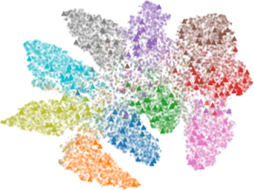}
      \caption*{(a) Round 4}
  \end{minipage}\\[2ex]
  \begin{minipage}[t]{0.23\textwidth}
    \centering
    \includegraphics[width=\textwidth]{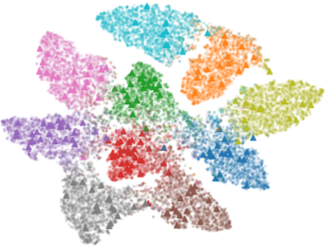}
       \caption*{(a) Round 6}
  \end{minipage}\hfill
  \begin{minipage}[t]{0.23\textwidth}
    \centering
    
    \includegraphics[width=\textwidth]{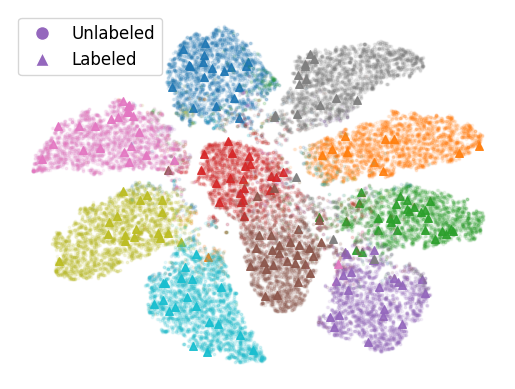}
      \caption*{(a) Round 8}
  \end{minipage}
  \caption{t-SNE visualization of feature embeddings generated by LPLgrad on the CIFAR-10 dataset. \vspace{-0.2cm}}
  \label{fig:Cluster}
\end{figure}


{\bf Evaluating LPLgrad's accuracy on a challenging dataset.} We evaluate LPLgrad on a more challenging dataset, CDD, in \fref{fig:disas} to demonstrate its effectiveness even with difficult datasets compared to our baselines. In \fref{fig:disas}(a), which depicts results with a high budget $\mathsf{A}$, LPLgrad starts with an accuracy of around $64\%$. By the first AL round, it surpasses all other baselines and continues to achieve the best performance across all rounds, reaching close to $75\%$. LearningLoss also shows strong performance, ending as the second-best performing algorithm. Ent-GradNorm attempts to catch up for almost half the rounds but then its performance degrades significantly. A similar trend is seen with entropy. Coreset and Rand both start well, but their performance declines over the rounds, with Coreset ending up performing worse than Rand.

\fref{fig:disas}(b) presents the results of CDD classification in a low-budget regime. Here, LPLgrad consistently outperforms other methods throughout all AL rounds, achieving $69\%$ accuracy. Due to the low-budget regime, the overall scale of accuracy is lower compared to the high annotation budget scenario. Entropy performs fairly well, securing the second-best position. Interestingly, Ent-GradNorm ends up last in this case. Coreset, random, and entropy continue to alternate their positions until the final round.

 \begin{figure}[!t]
  \centering
  \begin{minipage}[t]{0.24\textwidth}
    \centering
    \includegraphics[width=\textwidth]{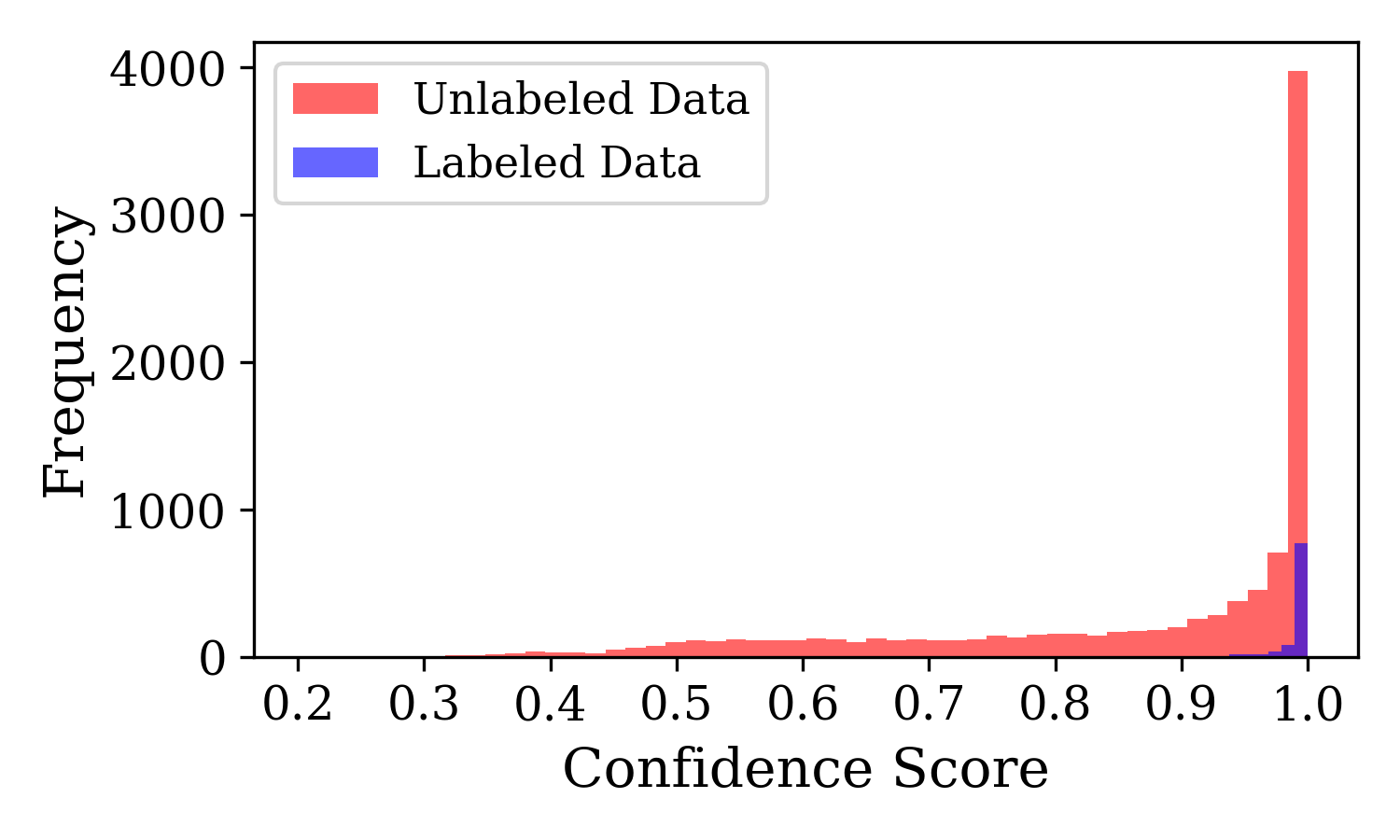}
    (a) Round 2
  \end{minipage}\hfill
  \begin{minipage}[t]{0.24\textwidth}
    \centering
    \includegraphics[width=\textwidth]{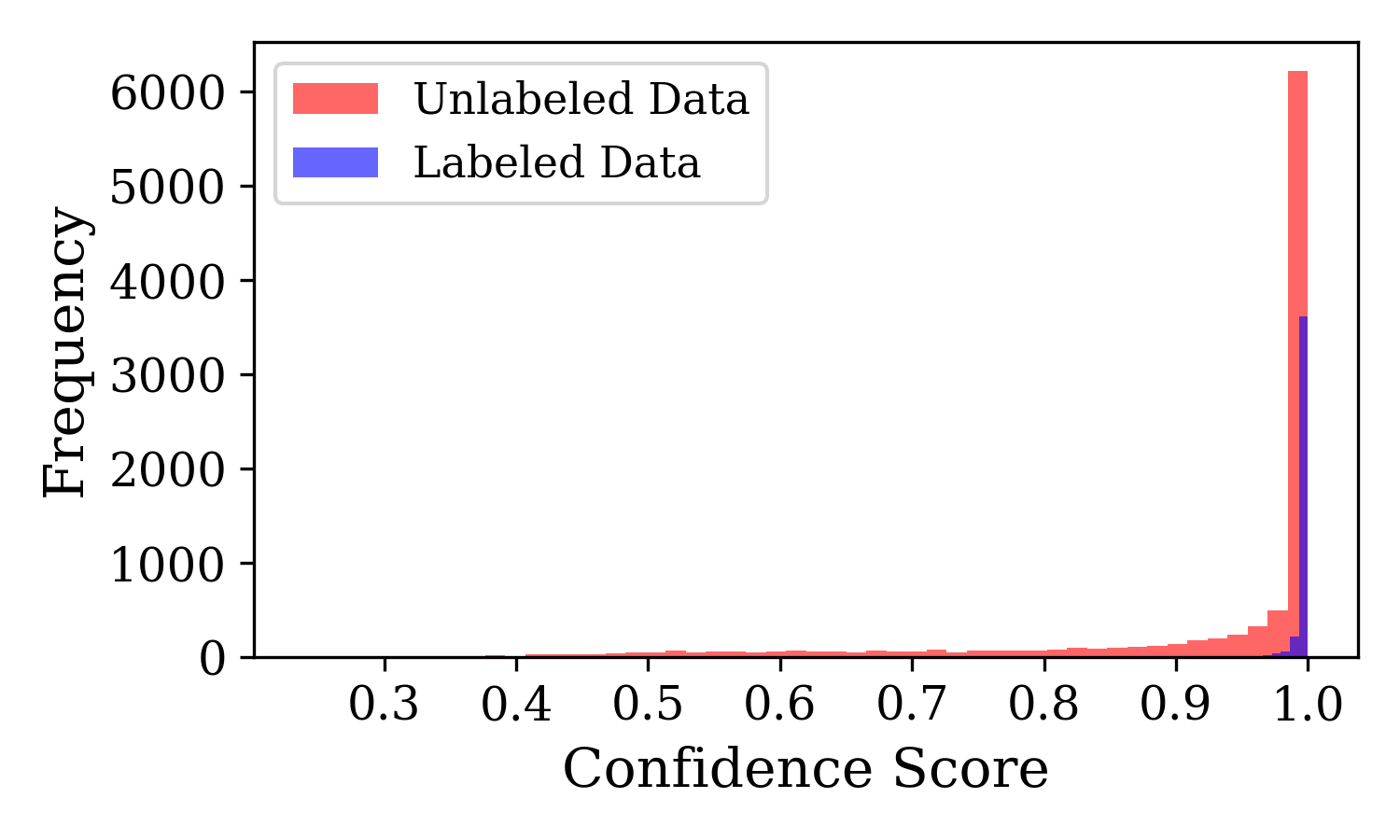}
    (b) Round 4
  \end{minipage}\\[2ex]
  \begin{minipage}[t]{0.24\textwidth}
    \centering
    \includegraphics[width=\textwidth]{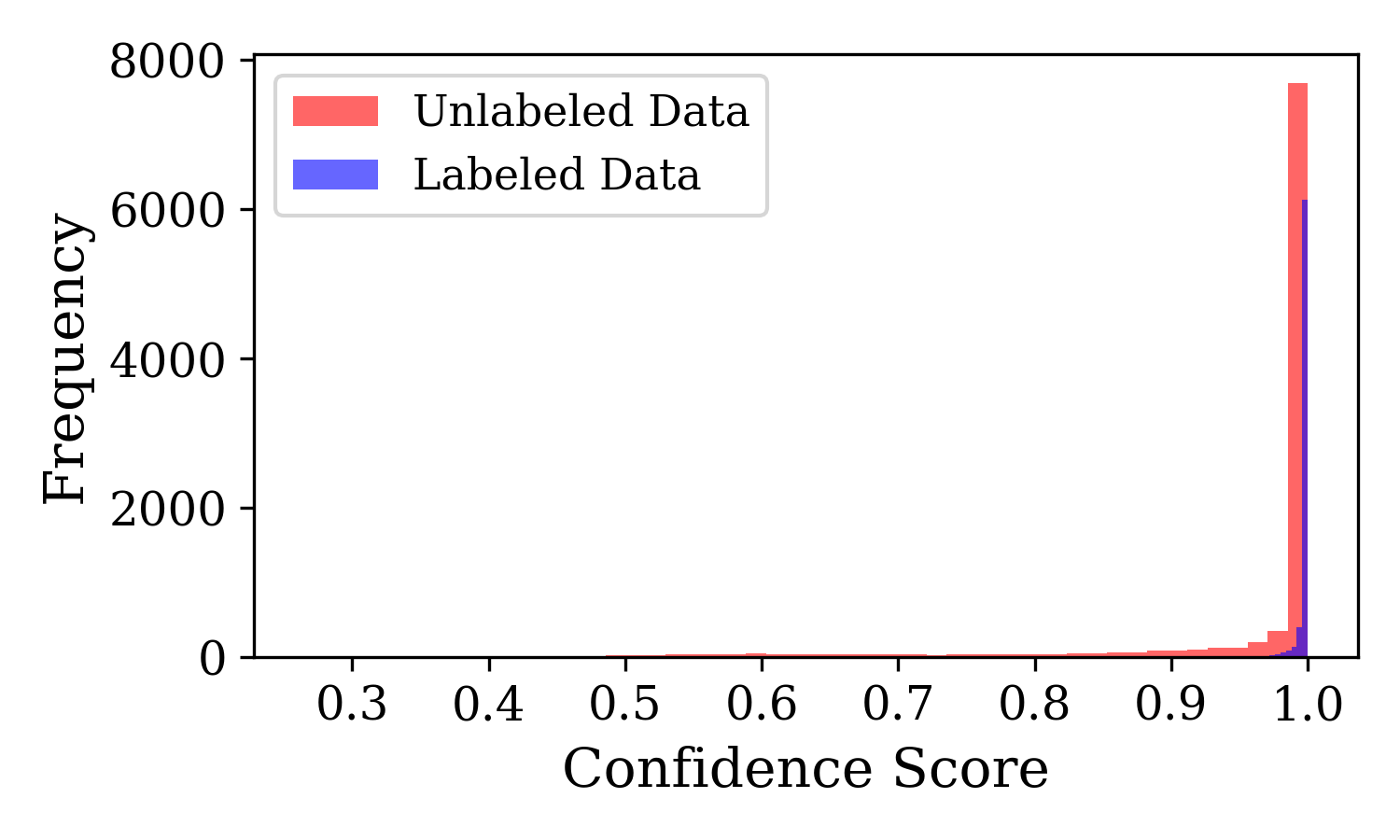}
    (c) Round 6
  \end{minipage}\hfill
  \begin{minipage}[t]{0.24\textwidth}
    \centering
    
    \includegraphics[width=\textwidth]{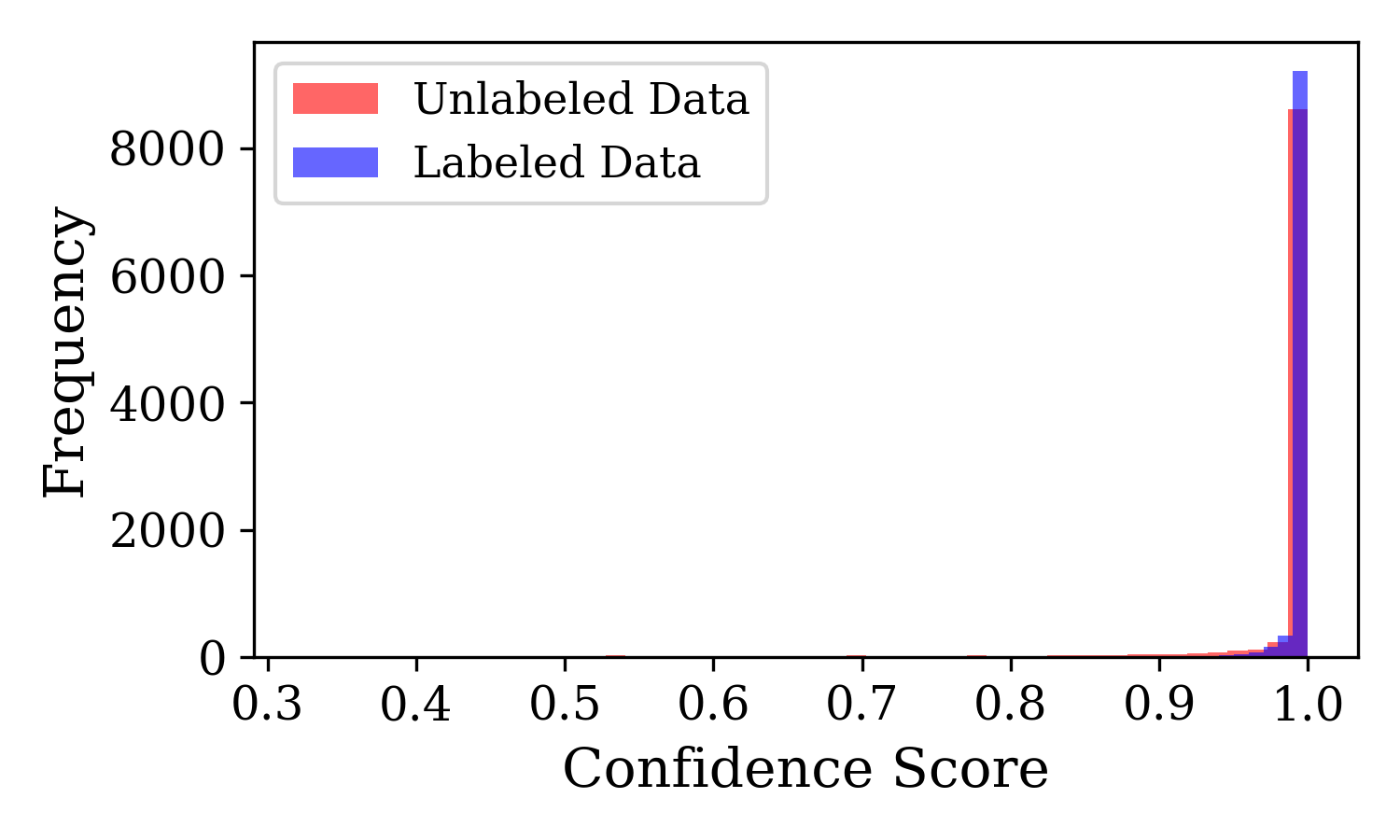}
    (d) Round 8
  \end{minipage}
  \caption{Confidence scores of $\boldsymbol{w}_{\text{main}}$ across various AL rounds on the CIFAR-10 dataset.\vspace{-0.2cm}}
  \label{fig:scores}
\end{figure}
{\bf Comparing LPLgrad's Computing time against baselines.} In tabel~\ref{table:performance_comparison}, we show the computation time required during the training and querying phases of LPLgrad against our baselines (using Nvidia A100 80GB RAM GPUs). Our observations indicate that Coreset is the most time-consuming algorithm due to its need to store all the data and compare new points with this stored data. On the other hand, LPLgrad performs comparably or even better than Entropy, Ent-GradNorm, and LearningLoss. While in some cases, LPLgrad may require more querying time due to backpropagation and parameter updates, this trade-off leads to improved accuracy, rendering the additional time investment worthwhile.

{\bf Evaluating LPLgrad's feature embeddings.} We first present in \fref{fig:Cluster} the t-distributed stochastic neighbor embedding (t-SNE) visualization of LPLgrad to illustrate the distribution of input features across four distinct AL rounds (2, 4, 6, 8). The visualization notably demonstrates that, {\em after only 8 rounds}, LPLgrad effectively labels CIFAR-10 images into their respective 10 classes. LPLgrad exhibits not only a strong discriminative ability in differentiating between instances of various classes but also shows enhanced proficiency in accurately assigning labels to these instances as the AL rounds progress.

{\bf Evaluating LPLgrad's Confidence score.} Next, in \fref{fig:scores}, we present the frequency distribution of the model’s confidence scores for labeled and unlabeled samples of CIFAR-10 dataset, particularly those close to 0.99, across the same AL rounds (2, 4, 6, 8). We plot the confidence scores for 10,000 unlabeled images in all the subfigures, while the number of labeled images increases by 1,000 in each subsequent graph, starting from 1,000. Specifically, \fref{fig:scores}(a) illustrates that the confidence scores for the unlabeled samples are more widely distributed across the confidence intervals. There are approximately 4,000 images for which the model is confident in their predictions, but there are 1,000 images with a confidence frequency of less than 0.99. In contrast, for the labeled images, almost 700 out of 1,000 have a confidence score between 0.99 and 1. \fref{fig:scores}(b) shows the frequency of scores for the fourth AL round. Evidently, 6,000 unlabeled images now have confidence scores between 0.99 and 1. Similarly, about 3,500 out of 4,000 are predicted correctly for the labeled samples, hence decreasing the spread of both the labeled and unlabeled samples’ confidence scores. Moving to \fref{fig:scores}(c), we observe a further decrease in the spread of bins, indicating that the model is more confident with both the labeled and unlabeled images. It gives scores between 0.9 and 1 for about 7,500 unlabeled and 6,000 labeled images. In \fref{fig:scores}(d), we show that the labeled images have surpassed the unlabeled ones, with about 8,500 and 8,200 out of 10,000, respectively, having scores between 0.99 and 1.

\begin{figure}[!t]
  \centering
  \begin{minipage}[!t]{0.5\linewidth}
    \centering
    \begin{adjustbox}{valign=t}
      \includegraphics[width=\linewidth]{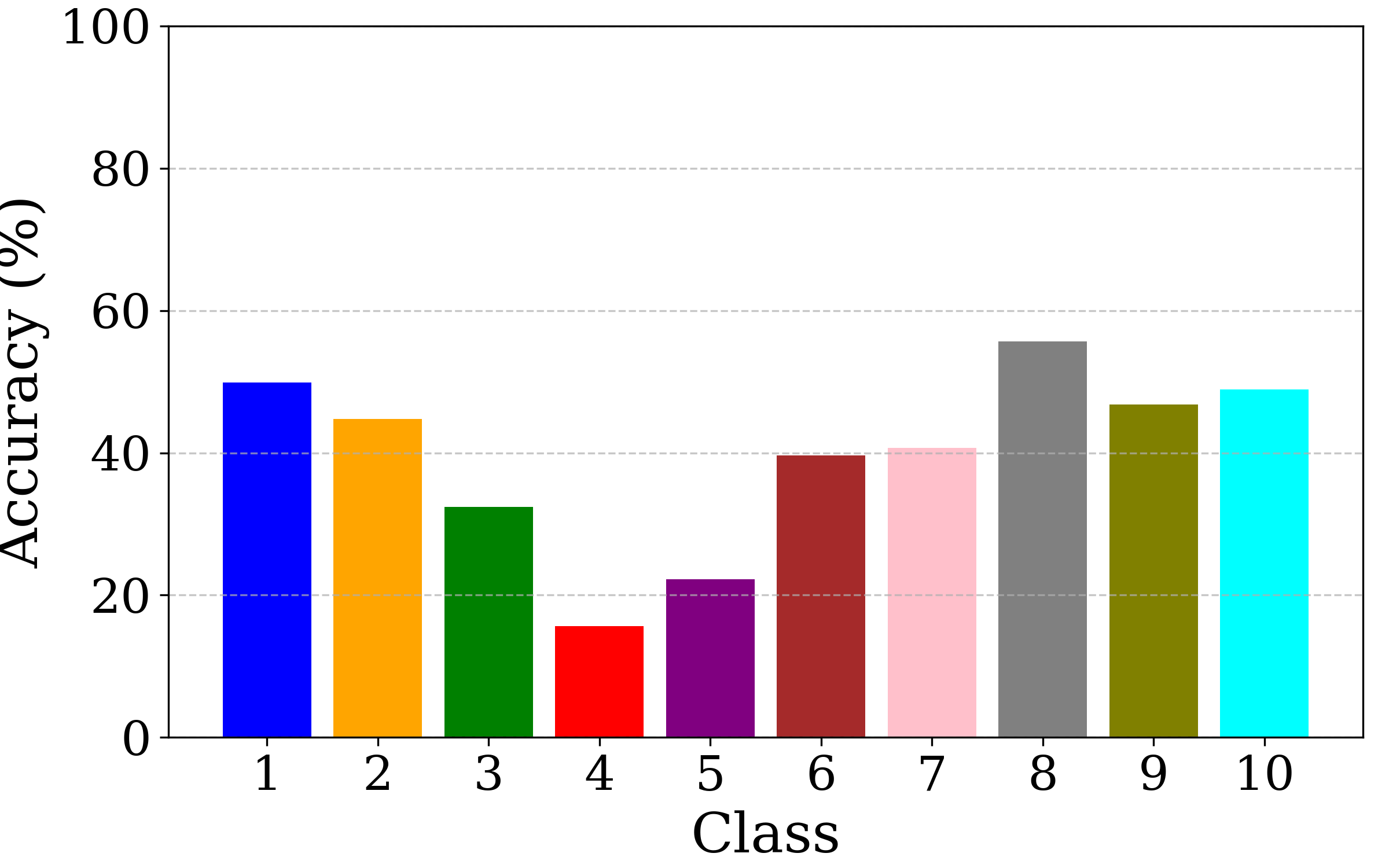}
    \end{adjustbox}
    \caption*{(a) Round 2}
  \end{minipage}\hfill
  \begin{minipage}[!t]{0.5\linewidth}
    \centering
    \begin{adjustbox}{valign=t}
      \includegraphics[width=\linewidth]{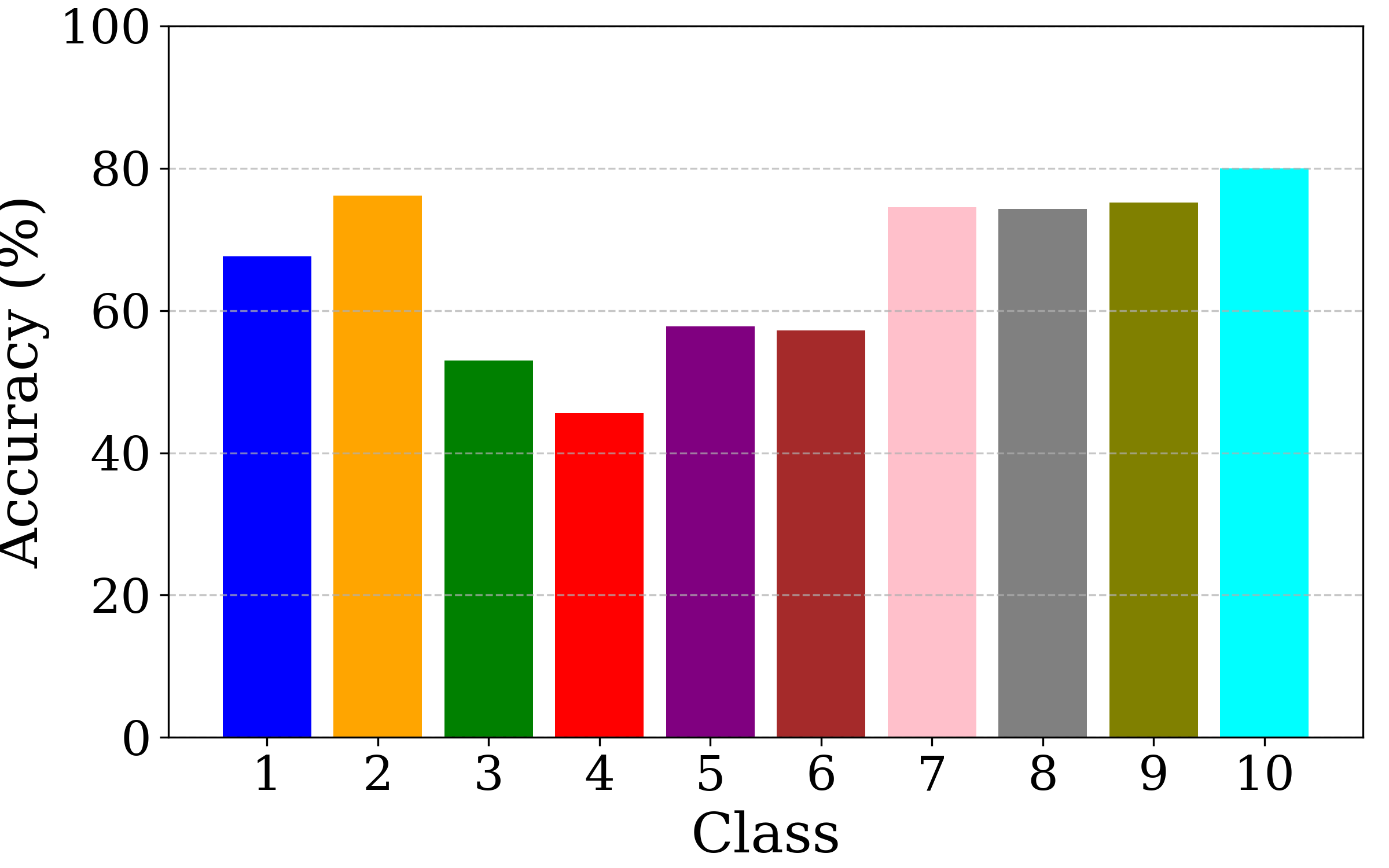}
    \end{adjustbox}
    \caption*{(b) Round 4}
  \end{minipage}\\[2ex]
  \begin{minipage}[!t]{0.5\linewidth}
    \centering
    \begin{adjustbox}{valign=t}
      \includegraphics[width=\linewidth]{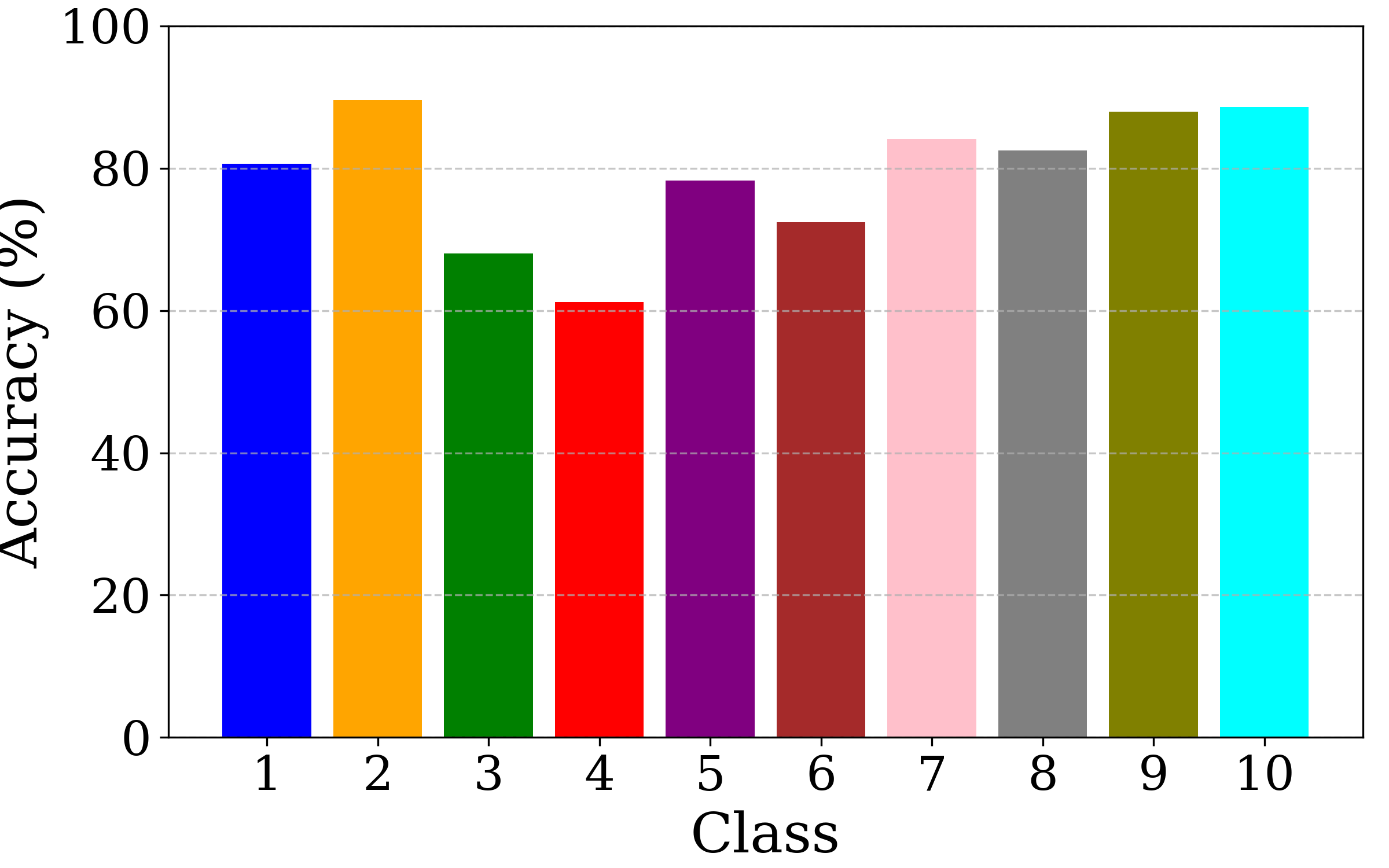}
    \end{adjustbox}
    \caption*{(c) Round 6}
  \end{minipage}\hfill
  \begin{minipage}[!t]{0.5\linewidth}
    \centering
    \begin{adjustbox}{valign=t}
      \includegraphics[width=\linewidth]{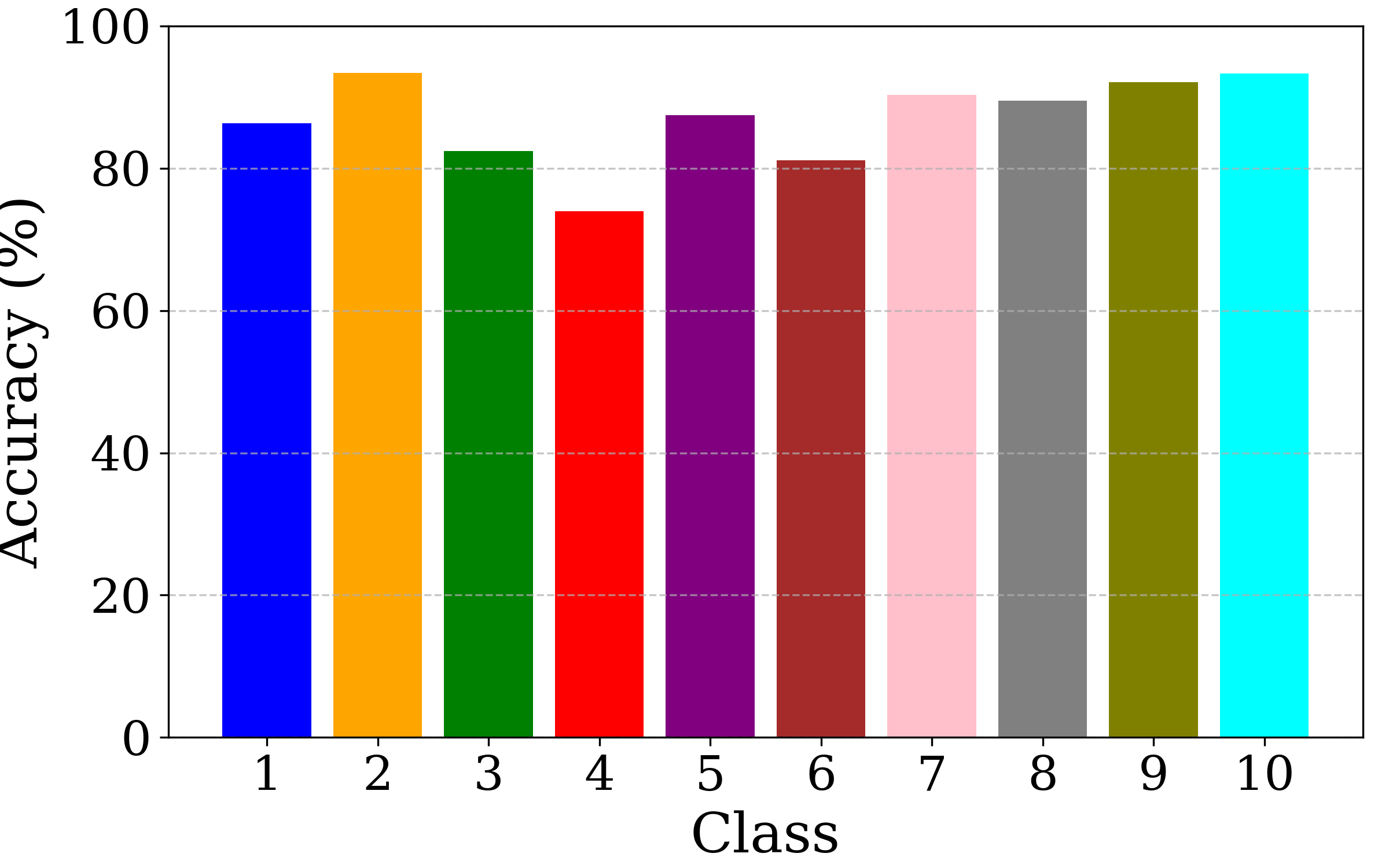}
    \end{adjustbox}
    \caption*{(d) Round 8}
  \end{minipage}
  \caption{Per class accuracy values of LPLgrad on CIFAR10. }
  \label{fig:per_class}
\end{figure}

{\bf Evaluating LPLgrad's class accuracy.} In \fref{fig:per_class}, we analyze the accuracy of LPLgrad across all 10 classes in the CIFAR-10 dataset across the four AL classes. Specifically,  \fref{fig:per_class} shows the results of the second AL round, where the model exhibits a lack of confidence, with the highest accuracy reaching only $55.7\%$. Moving to \fref{fig:per_class}(b), which depicts the fourth AL round, we observe significant improvements across all classes, with the model achieving a peak accuracy of $80\%$. \fref{fig:per_class}(c) illustrates the sixth AL round, where the model's proficiency continues to improve, achieving the highest accuracy of $88.6\%$. Finally, \fref{fig:per_class}(d) shows the results of the eighth AL round, where the model demonstrates a thorough understanding of the dataset, achieving consistently high accuracies across all classes, with the highest accuracy reaching 93.3\%.


\subsection{Ablation Study}

In this subsection, we delve into the significance of incorporating an auxiliary model to enhance the accuracy of our training pipeline. We explore the impact of integrating this auxiliary model by comparing results obtained from standalone training (only using the main model $w_{main}$) with those achieved when the auxiliary model is used. The primary role of the auxiliary model is to bolster the main model's capacity to learn and process input features more efficiently. It achieves this by introducing an additional set of layers that are trained concurrently with the main model. This setup allows the auxiliary model to capture supplementary information and nuances from the data, which the main model alone might overlook. By doing so, we aim to improve the overall performance, robustness, and generalization of the main model. The following sections will present detailed results and analyses of these approaches to highlight the tangible benefits of including the auxiliary model in our training regimen.

In Table~\ref{tab:low_abl} and~\ref{tab:high_abl}, we present a comparison under two different annotation budget settings, $\mathsf{A}$, across four distinct datasets. These results illustrate that the auxiliary model consistently supports the main model in achieving higher performance throughout the entire training process. Specifically, the inclusion of the auxiliary model leads to an approximate improvement in accuracy of around 8\%. This enhancement is observed across all training epochs, demonstrating the auxiliary model's effectiveness in improving the main model's performance and stability. The tables provide detailed evidence of the auxiliary model's positive impact on both accuracy and result consistency, regardless of the training phase.


\section{Conclusion and Future Work}
In this paper, we proposed LPLgrad, a novel AL approach aimed at addressing a common gap in the literature: the underutilization of the core phases of AL, specifically the training and querying phases. To fully exploit the labeled data, we adopted an augmented approach wherein two models—a main model and an auxiliary model—are trained together to optimally learn the features of the input data. Additionally, to effectively query the most informative samples, we compute the entropy values of the unlabeled set and backpropagate these values to obtain loss values. This loss is minimized, and its gradients with respect to the main model's parameters are calculated. The Frobenius norm of gradients is then computed, sorted, and used to identify the samples with the highest gradient norm values, which are selected for labeling and added to the labeled set. We extensively evaluate LPLgrad on diverse image classification datasets and a real-world dataset to validate its efficacy. Our findings demonstrate that LPLgrad surpasses state-of-the-art approaches by achieving higher accuracy with fewer labels and less computing time.

As future work, we plan to explore the application of LPLgrad to more complex and larger-scale datasets to further validate its robustness and scalability. We also aim to investigate the potential of LPLgrad in domains beyond image classification, such as natural language processing or time-series data, which will be a key focus. Finally, we will explore using LPLgrad in real-time systems where rapid decision-making with limited labeled data is critical.

\begin{table}[!t]
\centering
\caption{Accuracy values across different epochs with and without the {auxiliary model}  model for {\bf low} $\mathsf{A}$.}
\label{tab:low_abl}
\setlength{\tabcolsep}{1em}
\renewcommand{\arraystretch}{1.2}
\resizebox{1\linewidth}{!}{
\begin{tabular}{p{1.4cm}| p{0.95cm}|p{0.95cm}|p{1cm}|p{0.95cm}|p{0.95cm}|p{1cm}}
\toprule
\multirow{2}{*}{} & \multicolumn{3}{c|}{with $\boldsymbol{w}_{\text{aux}}$} & \multicolumn{3}{c}{w/o $\boldsymbol{w}_{\text{aux}}$} \\ \cline{2-7} 
                  & 1/3 of A & 1/2 of A & Entire A & 1/3 of A & 1/2 of A & Entire A \\ \hline
CIFAR-10          & 52.1    & 65.7     &  77.4  & 49.8     & 63.3     & 76.2   \\ \hline
CIFAR-100         & 24.6     &  43.7    &  50.2  &  24.1    &   42.3   & 47.7   \\ \hline
SVHN              & 75.2     &  88.7    & 92.4   & 69.7     & 86.3     &   91.2 \\ \hline
Caltech-101       & 22.5     & 31.4     & 37.6   & 14.9     &27.6      & 34.8   \\ 
\bottomrule
\end{tabular}}
\end{table}
\begin{table}[!t]
\centering
\setlength{\tabcolsep}{1em}
\renewcommand{\arraystretch}{1.2}
\caption{Accuracy values across different epochs with and without the {auxiliary model} for {\bf high} $\mathsf{A}$.}
\label{tab:high_abl}
\resizebox{1\linewidth}{!}{
\begin{tabular}{p{1.4cm}| p{0.95cm}|p{0.95cm}|p{1cm}|p{0.95cm}|p{0.95cm}|p{1cm}}
\toprule
\multirow{2}{*}{} & \multicolumn{3}{c|}{with $\boldsymbol{w}_{\text{aux}}$} & \multicolumn{3}{c}{w/o $\boldsymbol{w}_{\text{aux}}$} \\ \cline{2-7} 
                  & 1/3 of A & 1/2 of A & Entire A & 1/3 of A & 1/2 of A & Entire A \\ \hline
CIFAR-10          & 72.2   & 85.5     &   90.8  & 69.8     & 83.7     & 89.9  \\ \hline
CIFAR-100         & 47.7     &  63.4   &  68.7  & 47.6    &   62.1   & 67.9  \\ \hline
SVHN              & 91     &  93.5    & 95.1   & 89.8     & 93.3     &   95 \\ \hline
Caltech-101       & 33.5    & 44.2     & 53.8  & 33.4     &43.8      &  51.2   \\ 
\bottomrule
\end{tabular}}
\end{table}

\bibliographystyle{IEEEtran}
{
\bibliography{ref.bib}
}
\end{document}